\documentclass[10pt,twocolumn,letterpaper]{article}

\usepackage{iccv}
\usepackage{times}
\usepackage{epsfig}
\usepackage{graphicx}
\usepackage{amsmath}
\usepackage{amssymb}
\usepackage{booktabs}
\usepackage{multirow}
\usepackage[caption=false]{subfig}

\usepackage[pagebackref=true,breaklinks=true,colorlinks,bookmarks=false]{hyperref}

\iccvfinalcopy 
\newcommand{\censored}[1]{}
\newcommand{\parspace}{\vspace{-3mm}}

\def\iccvPaperID{000} 

\ificcvfinal\fi
\ificcvfinal\renewcommand{\censored}[1]{#1}\fi

\begin{document}
\title{KNEEL: Knee Anatomical Landmark Localization Using Hourglass Networks}

\author{Aleksei Tiulpin\\
University of Oulu,\\
Oulu University Hospital\\
{\tt\small aleksei.tiulpin@oulu.fi}
\and
Iaroslav Melekhov\\
Aalto University\\
{\tt\small iaroslav.melekhov@aalto.fi}
\and
Simo Saarakkala\\
University of Oulu,\\
Oulu University Hospital\\
{\tt\small simo.saaarakkala@oulu.fi}
}

\maketitle
\begin{abstract}
This paper addresses the challenge of localization of anatomical landmarks in knee X-ray images at different stages of osteoarthritis (OA). Landmark localization can be viewed as regression problem, where the landmark position is directly predicted by using the region of interest or even full-size images leading to large memory footprint, especially in case of high resolution medical images. In this work, we propose an efficient deep neural networks framework with an hourglass architecture utilizing a soft-argmax layer to directly predict normalized coordinates of the landmark points. We provide an extensive evaluation of different regularization techniques and various loss functions to understand their influence on the localization performance. Furthermore, we introduce the concept of transfer learning from low-budget annotations, and experimentally demonstrate that such approach is improving the accuracy of landmark localization. Compared to the prior methods, we validate our model on two datasets that are independent from the train data and assess the performance of the method for different stages of OA severity. The proposed approach demonstrates better generalization performance compared to the current state-of-the-art.
\end{abstract}

\section{Introduction}
Anatomical landmark localization is a challenging problem that appears in many medical image analysis problems~\cite{payer2019integrating}. One particular realm where the localization of landmarks is of high importance is the analysis of knee plain radiographs at different stages of osteoarthritis (OA) -- the most common joint disorder and $11^{th}$ highest disability factor in the world~\cite{allen2015epidemiology}.

\begin{figure}[t!]
\centering
\includegraphics[width=0.9\linewidth]{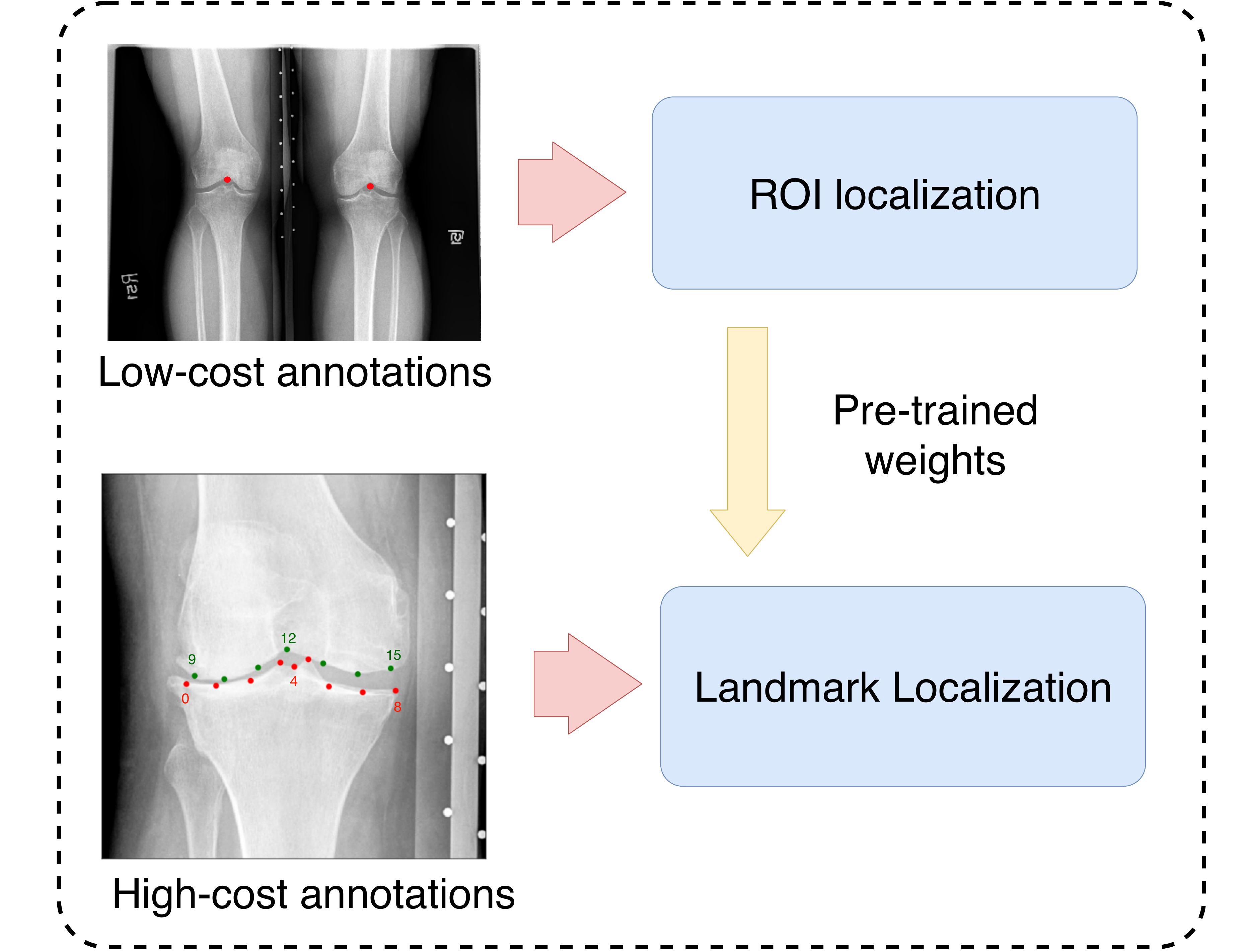}
\caption{Graphical illustration of our approach. At the first stage, the knee joint area localization model is trained using low-cost annotations. At the second stage, we leverage the weights of the model pre-trained using the low-cost annotations and train a model that localizes $16$ individual landmarks. The numbers in the figure indicate the landmark ID (best viewed on screen). The tibial landmarks are displayed in red and numbered from $0$ to $8$ (left-to-right). Femoral landmarks are displayed in green and numbered from $9$ to $15$ (left-to-right). }\label{fig:pipeline}
\end{figure}

In knee OA research field, as well as in the other domains, two sub-tasks that form a typical pipeline for landmark localization can be defined: the region of interest (ROI) localization and the landmark localization itself~\cite{wu2019facial}. In knee radiographs, the former one is typically applied in the analysis of the whole knee images~\cite{antony2017automatic,antony2016quantifying,norman2019applying,tiulpin2019multimodal,tiulpin2018automatic}, while the latter is used for bone shape and texture analyses~\cite{brahim2019decision,janvier2015roi,thomson2015automated}. Furthermore, Tiulpin~\emph{et al.} also used the landmark localization for image standardization applied after the ROI localization step~\cite{tiulpin2019multimodal,tiulpin2019oarsigrading}.

\begin{figure*}
\hfill
\subfloat[KL 0]{\includegraphics[width=0.2\textwidth]{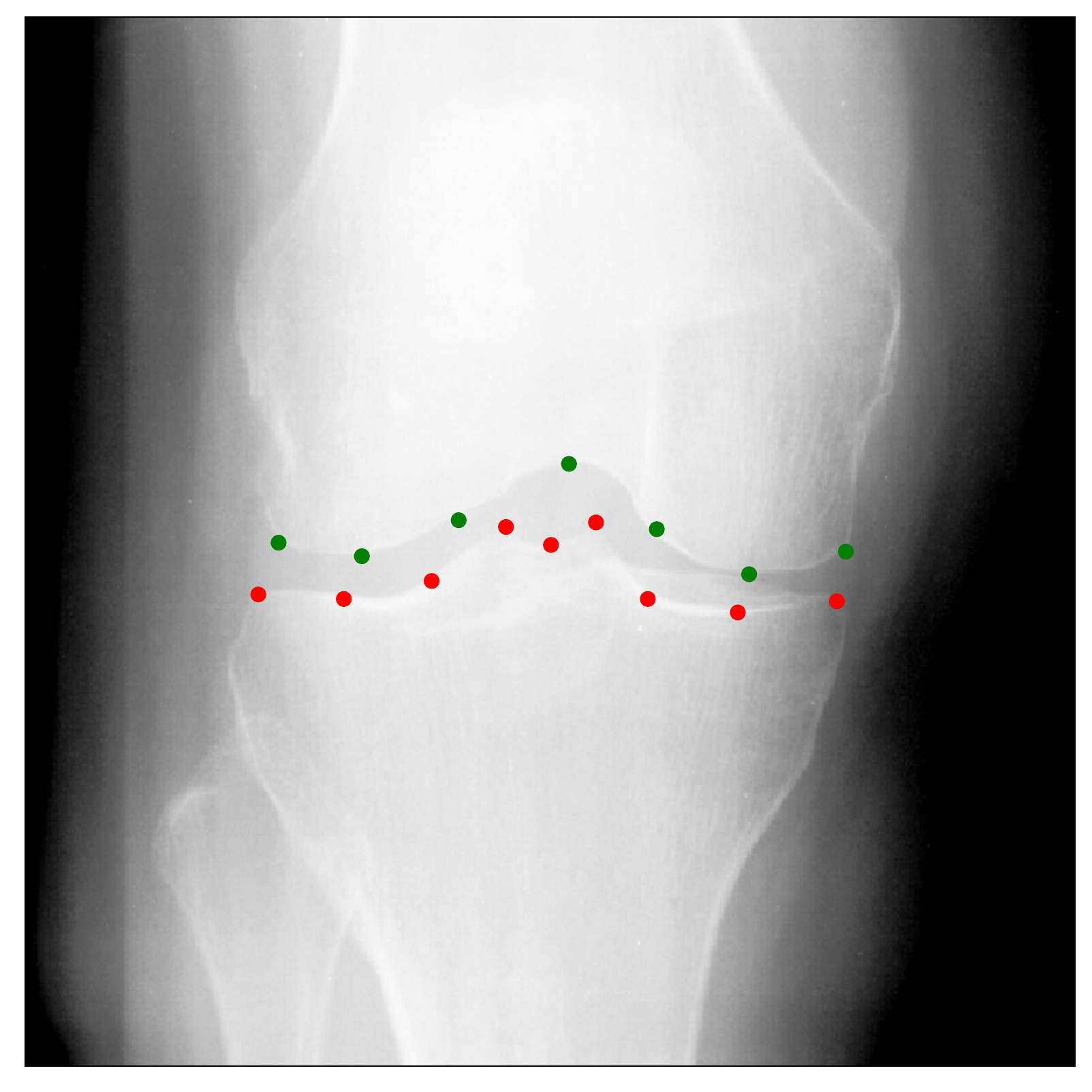}}
\subfloat[KL 1]{\includegraphics[width=0.2\textwidth]{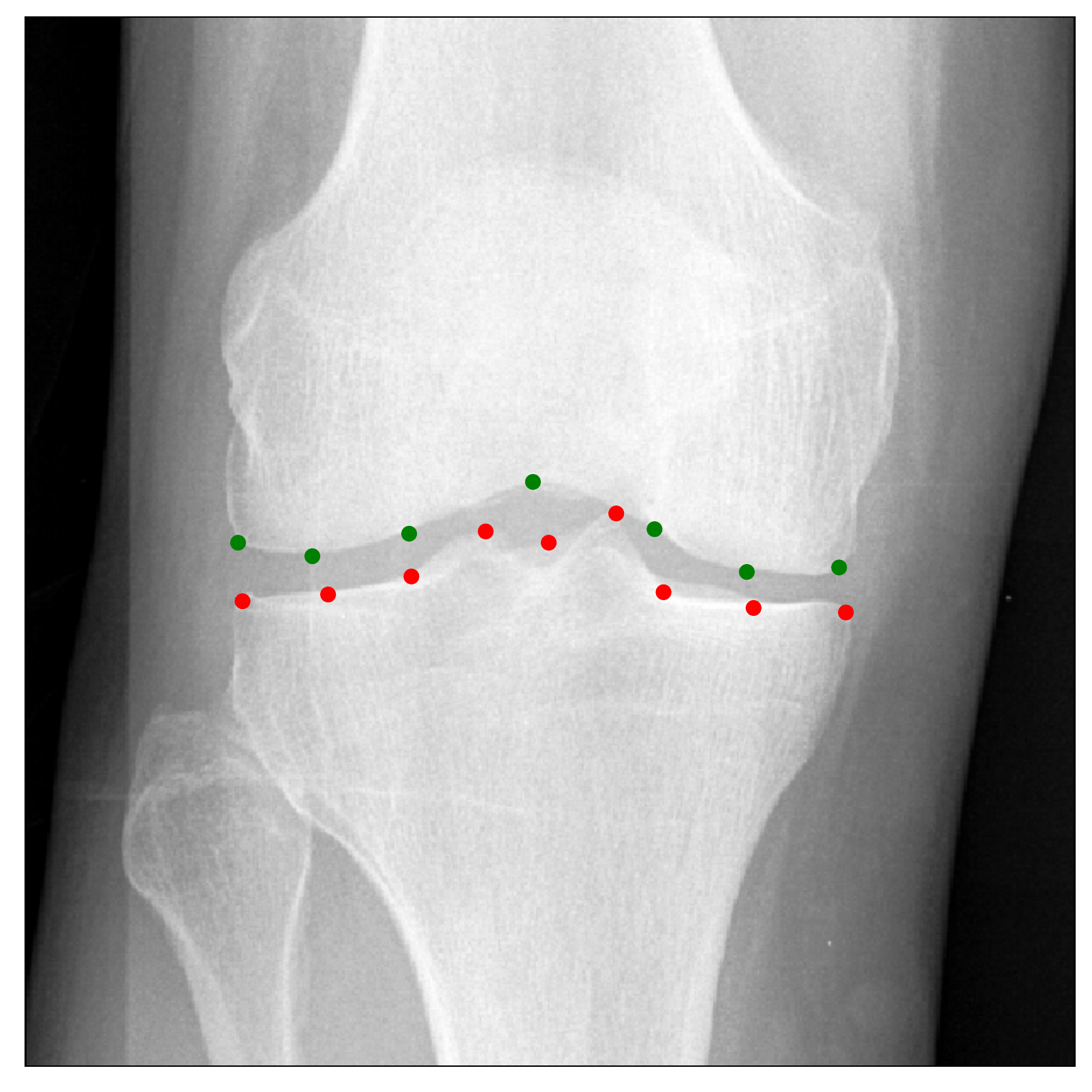}}
\subfloat[KL 2]{\includegraphics[width=0.2\textwidth]{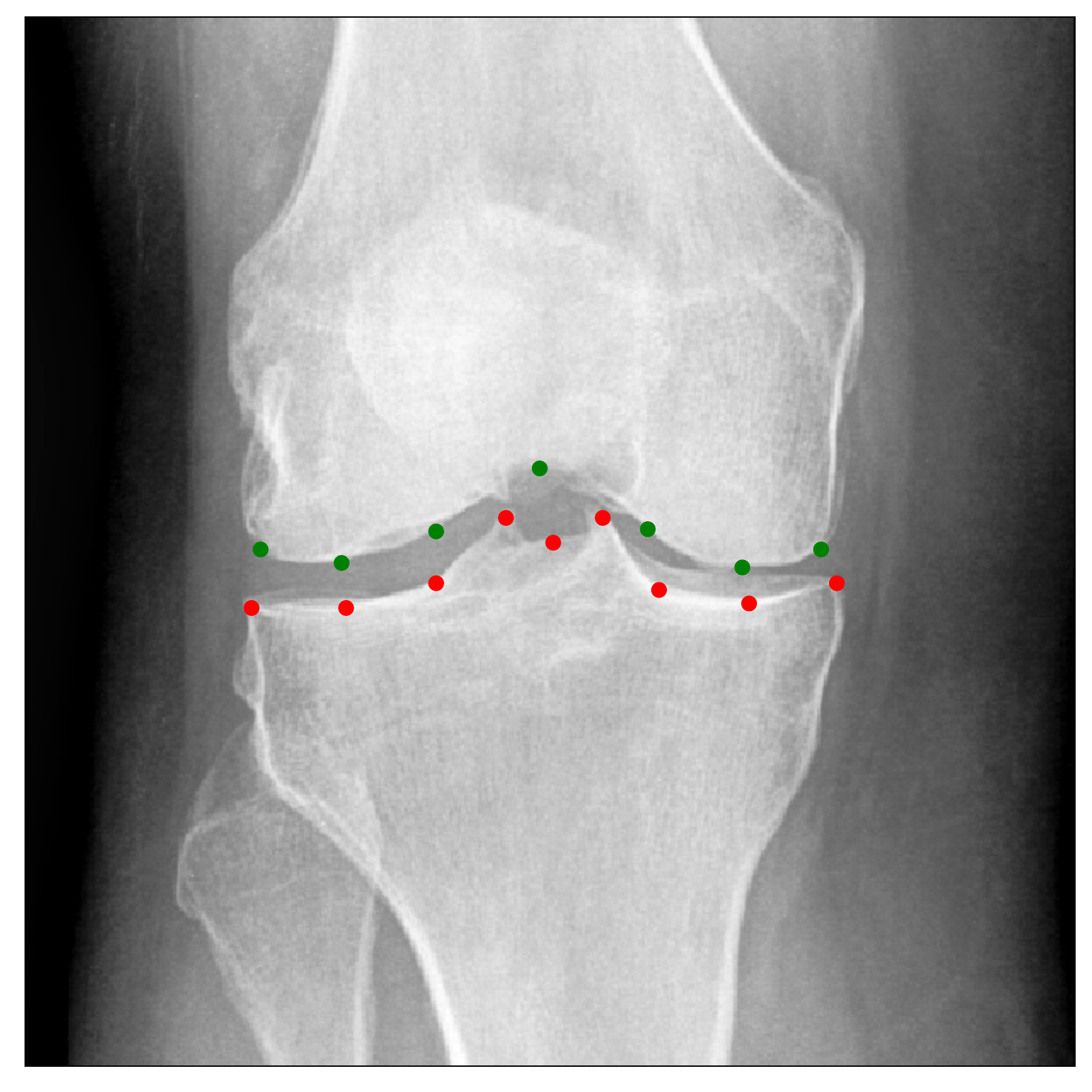}}
\subfloat[KL 3]{\includegraphics[width=0.2\textwidth]{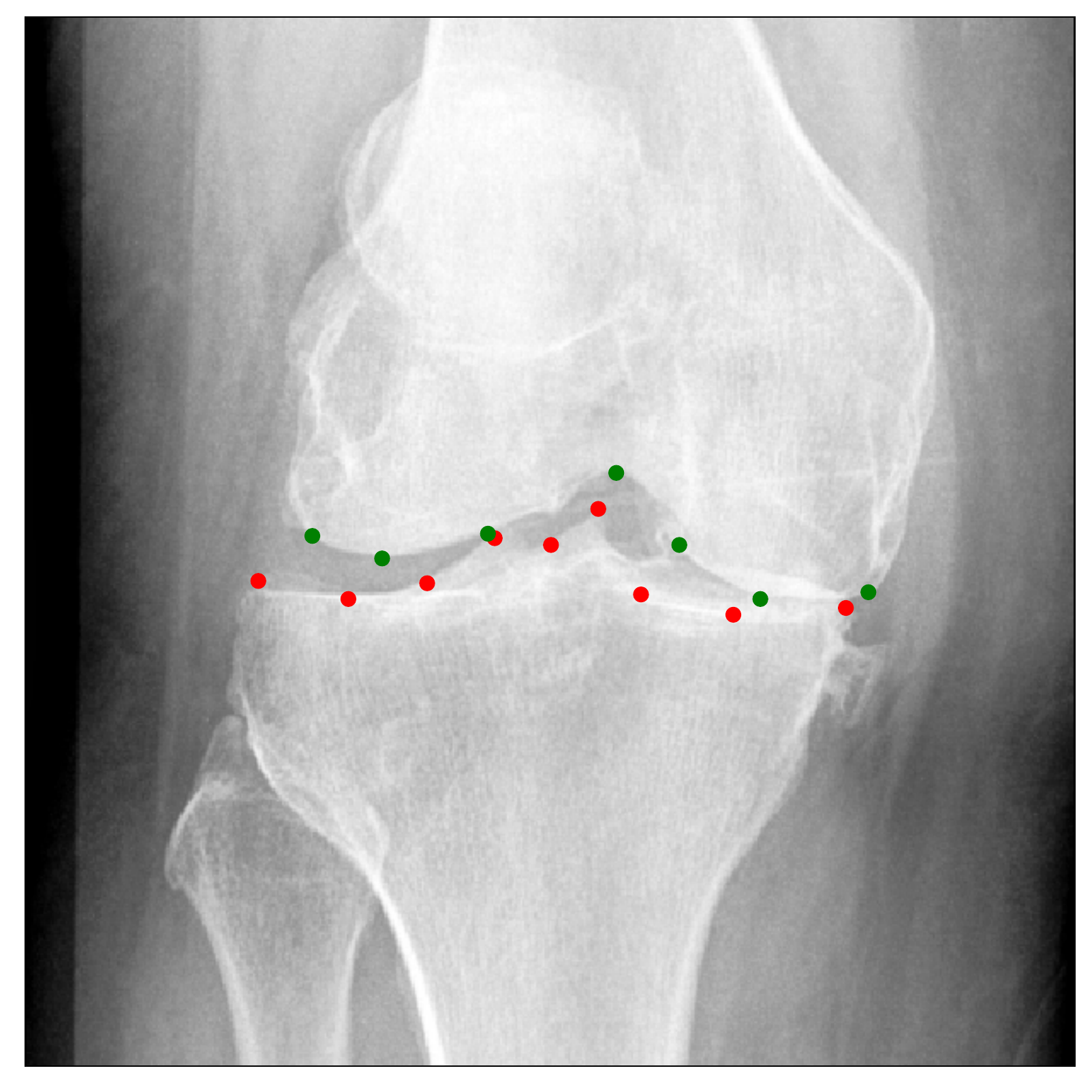}}
\subfloat[KL 4]{\includegraphics[width=0.2\textwidth]{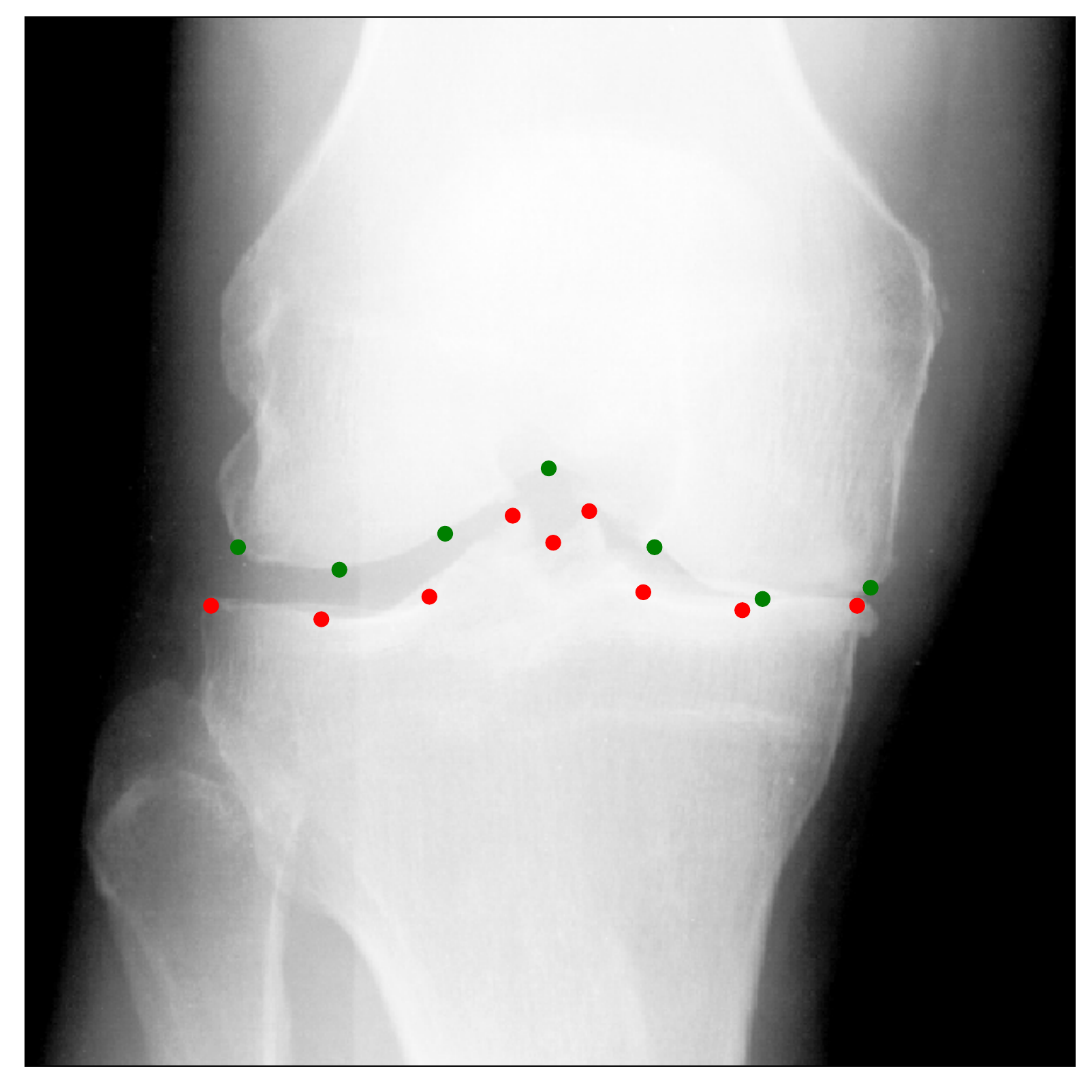}}
\hfill
\vspace{0.1cm}
\caption{Typical examples of knee joint radiographs at different stages of osteoarthritis severity with overlayed landmarks. Here, the images are cropped to $140\times140$ mm regions of interest. KL$\geq 2$ indicates radiographic osteoarthritis. This figure is best viewed on screen.}\label{fig:ann-examples}
\end{figure*}

Manual annotation of knee landmarks is not a trivial problem without the knowledge of knee anatomy, and it becomes even more challenging when the severity of OA increases. In particular, it makes the annotation process of fine-grained bone edges and tibial spines intractable and time consuming. In Fig.~\ref{fig:ann-examples}, we show the examples of annotations of the landmarks for each stage of OA severity graded according to the gold-standard Kellgren-Lawrence system (grading from $0$ to $4$)~\cite{kellgren1957radiological}. It can be seen from this figure that when the severity of the disease progresses, bone spurs (osteophytes) and the general bone deformity affect the appearance of the image. Other factors, such as X-ray beam angle are also known to have impact on the image appearance~\cite{kothari2004fixed}.

In this paper, we propose a novel Deep Learning based framework for localization of anatomical landmarks in knee plain radiographs and validate its generalization performance. First, we train a model to localize ROIs in a bilateral radiograph using low-cost labels, and subsequently, train a model on the localized ROIs to predict the location of $16$ anatomical landmarks in femur and tibia. Here, we utilize transfer learning and use the model weights from the first step of our pipeline for initialization of the second-stage model. The proposed approach is schematically illustrated in Fig.~\ref{fig:pipeline}.

Our method is based on the hourglass convolutional network~\cite{newell2016stacked} that localizes the landmarks in a weakly-supervised manner and subsequently uses the soft-argmax layer to directly estimate the location of every landmark point. To summarize, the contributions of this study are the following:

\begin{itemize}
    \item We leverage recent advances in landmark detection using hourglass networks and combine the best design choices in our method.
    \item For the first time, we propose to use MixUp~\cite{zhang2017mixup} data augmentation principle for anatomical landmark localization and perform a thorough ablation study for the knee radiographs.
    \item We demonstrate an effective strategy of enhancing the performance of our landmark localization method by pre-training it on low-budget landmark annotations.
    \item We evaluate our method on two independent datasets and demonstrate better generalization ability of the proposed approach compared to the current state-of-the-art baseline.
    \item The pre-trained models, source code and the annotations performed for the Osteoarthritis Initiative (OAI) dataset are publicly available at~\url{https://github.com/MIPT-Oulu/KNEEL}.
\end{itemize}

\section{Related Work}
In the literature, there exist only a few studies specifically focused on localization of landmarks in plain knee radiographs. Specifically, the current state-of-the-art was proposed by Lindner~\emph{et.al}~\cite{lindner2014robust,lindner2013accurate} and it is based on a combination of random forest regression voting (RFRV) with constrained local models (CLM) fitting.

There are several methods focusing solely on the ROI localization. Tiulpin~\emph{et al.}~\cite{tiulpin2017novel} proposed a novel anatomical proposal method to localize the knee joint area. Antony~\emph{et al.}~\cite{antony2017automatic} used fully convolutional networks for the same problem. Recently, Chen~\emph{et al.}~\cite{chen2019fully} proposed to use object detection methods to measure the knee OA severity.

The proposed approach is related to the regression-based methods for keypoint localization~\cite{wu2019facial}. We utilize an hourglass network which is an encoder-decoder model initially introduced for human pose estimation~\cite{newell2016stacked} and address both ROI and landmark localization tasks. Several other studies in medical imaging domain also leveraged a similar approach by applying U-Net~\cite{ronneberger2015u} to the landmark localization problem~\cite{davison2018landmark,payer2019integrating}. However, the encoder-decoder networks are computationally heavy during the training phase since they regress a tensor of high-resolution heatmaps which is challenging for medical images that are typically of a large size. It is notable that decreasing the image resolution could negatively impact the accuracy of landmark localization. In addition, most of the existing approaches use a refinement step which makes the computational burden even harder to cope with. Nevertheless, hourglass CNNs are widely used in human pose estimation~\cite{newell2016stacked} due to a possibility of lowering down the resolution and the absence of precise ground truth.

More similar to our approach, Honari~\emph{et al.}~\cite{honari2018improving} recently leveraged deep learning and applied soft-argmax layer to the feature maps of the full image resolution to improve landmark localization performance leading to remarkable results. However, such strategy is computationally heavy for medical images due to their high resolution. In contrast, we first moderately reduce the image resolution by embedding it into a feature space, utilize an hourglass module to process the obtained feature maps at all scales, and eventually apply the soft-argmax operator that makes the proposed configuration more applicable to high-resolution images allowing to get sub-pixel accurate landmark coordinates.

\section{Method}
\subsection{Network architecture}
\begin{figure}[ht!]
    \centering
    \subfloat[]{\includegraphics[height=0.3\textwidth]{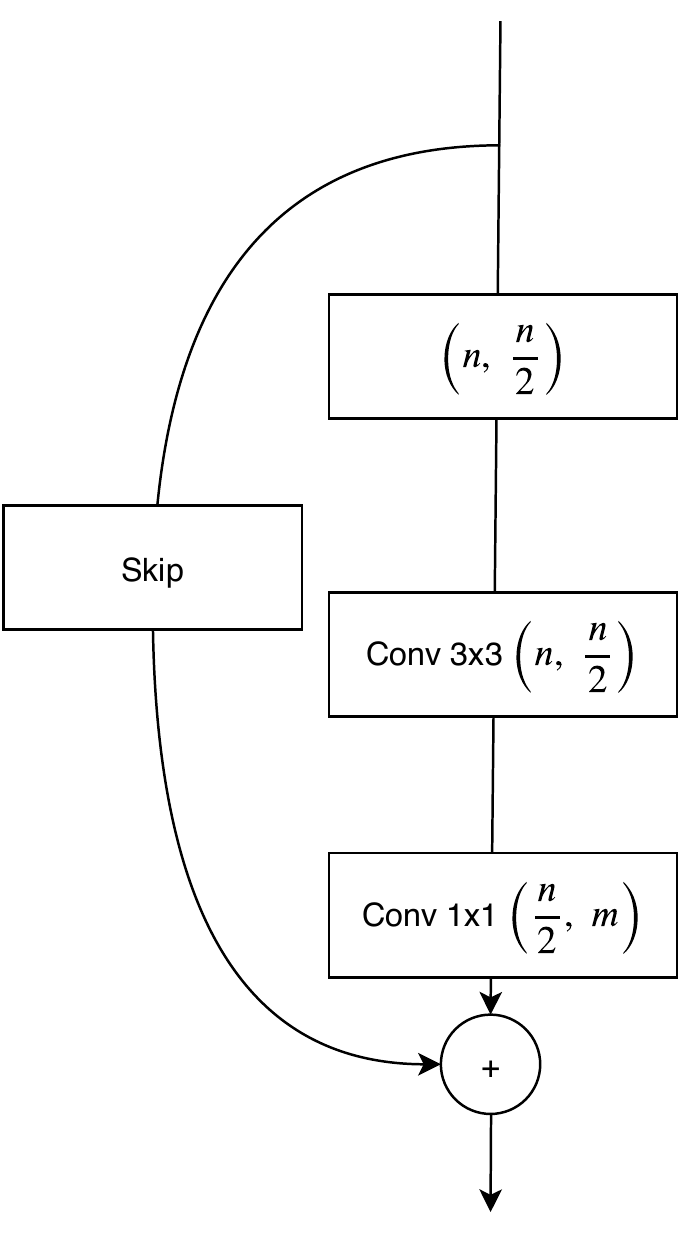}}\hfill
    \subfloat[]{\includegraphics[height=0.3\textwidth]{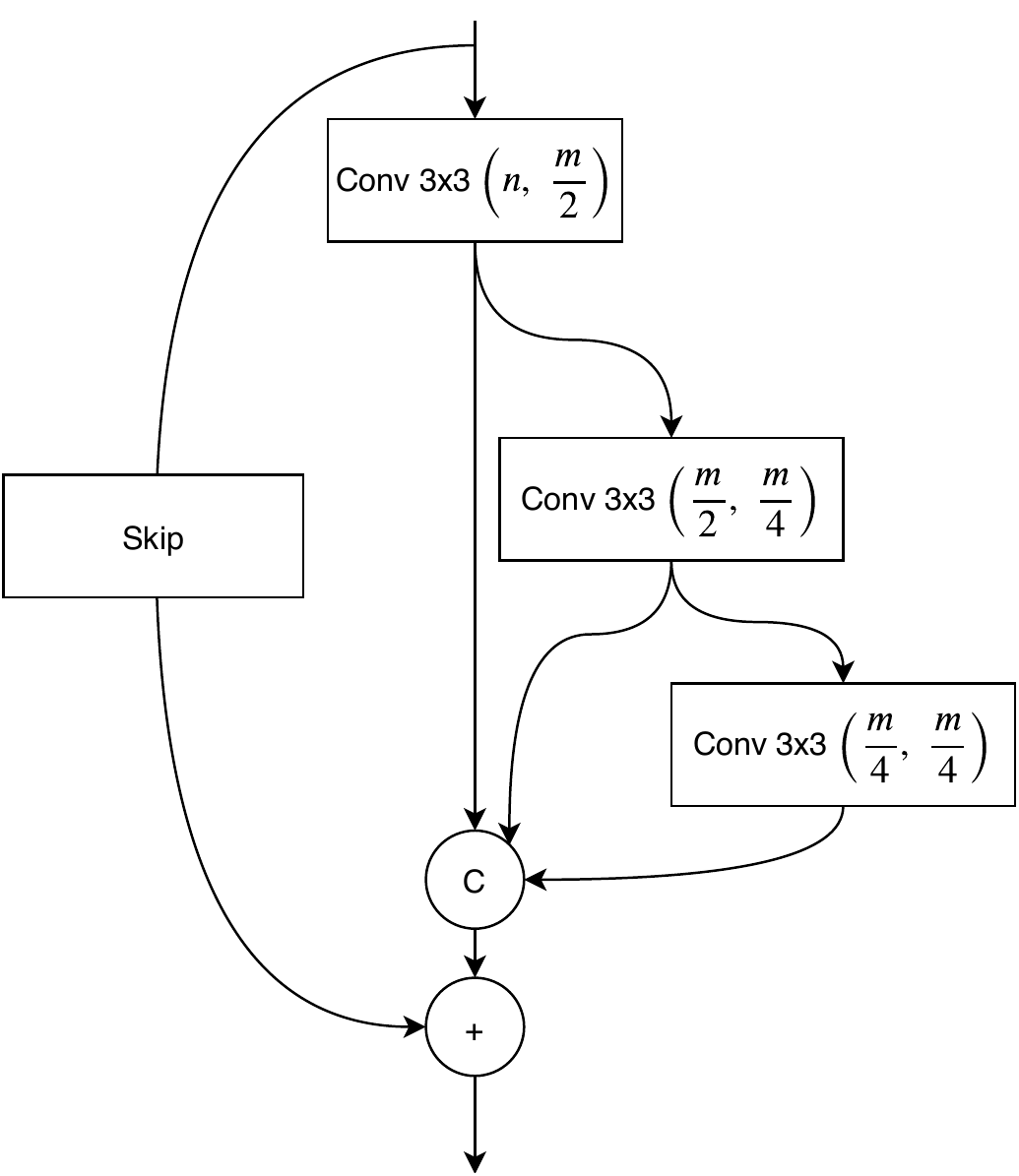}}
    \vspace{0.1cm}
    \caption{Graphical illustration of the difference between the bottleneck residual block~\cite{newell2016stacked,he2016deep} (a) and the HMP residual block~\cite{bulat2017binarized} (b). Here, $n$ and $m$ indicate the number of input and output feature maps, respectively. Skip connection representing $1\times1$ convolution is applied if $n \neq m $.}\label{fig:hg-modules}
\end{figure}

\paragraph{Overview.}
Our model comprises several architectural components of modern hourglass-like encoder-decoder models for landmark localization. In particular, we utilize the hierarchical multi-scale parallel (HMP) residual block~\cite{bulat2017binarized} which improves the gradient flow compared to the traditional bottleneck layer described in:~\cite{he2016deep,newell2016stacked}.  The HMP block structure is illustrated in~Fig.~\ref{fig:hg-modules}.

\begin{figure*}
    \centering
    \includegraphics[width=\linewidth]{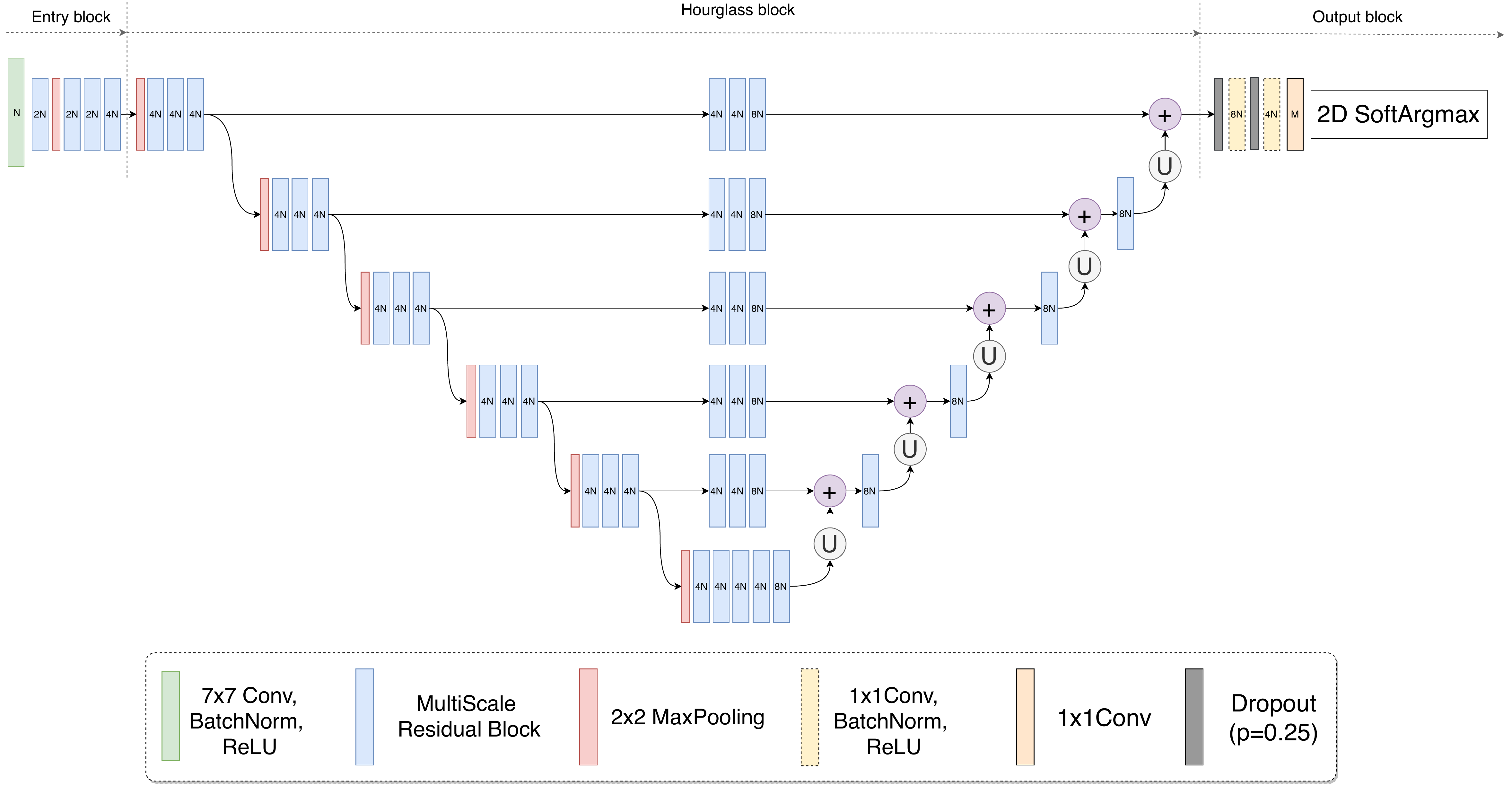}
    \vspace{0.1cm}
    \caption{Model architecture with an hourglass block of depth $d=6$. Here, $N$ is a width of the network and $M$ is the number of output landmarks.}
    \label{fig:arch}
\end{figure*}

The architecture of the proposed model is represented in~Fig.~\ref{fig:arch}. In general, our model comprises three main components: entry block, hourglass block, and output block. The whole network is parameterized by two hyperparameters -- width $N$ and depth $d$, where the latter is related to the number of max-pooling steps in the hourglass block. In our experiments we found the width of $N=24$ and the depth of $d=6$ to be optimal to maintain both high accuracy and speed of computations.

\parspace
\paragraph{Entry block.} Similar to the original hourglass model~\cite{newell2016stacked} we apply a $7\times 7$ convolution with stride $2$ and zero padding of $3$ and pass the results into a residual module. Further, we use a $2\times 2$ max-pooling and utilize three residual modules before the hourglass block. This block allows to simultaneously downscale the image $4$ times and obtain representative feature embeddings suitable for multi-scale processing performed in the hourglass block.

\parspace
\paragraph{Hourglass block.}
This block starts with a $2\times 2$ max-pooling and recursively repeats dual-path structure $d$ times as can be seen in Fig.~\ref{fig:arch}. In particular, each level of the hourglass block starts with a $2\times 2$ max-pooling subsequently followed by $3$ HMP residual blocks. At the next stage, the representations from the current level $i$ are passed to the next hourglass' level $i+1$ and also passed forward to be summed with the up-sampled outputs of the hourglass level $i+1$. Since spatial resolution of the feature maps at level $i$ and $i+1$ is different, the nearest-neighbours up-sampling is used~\cite{newell2016stacked}. At level $d$, we simply feed the representations into the HMP block instead of the next hourglass level due to the reached limit of hourglass' depth.

\parspace
\paragraph{Output block.}
The final block of the model uses the representations coming from the hourglass module and sequentially applies two blocks of dropout ($p=0.25$) and $1\times 1$ convolutional block with batch normalization and ReLU. At the final stage, a $1\times 1$ convolution and soft-argmax~\cite{chapelle2010gradient} are utilized to regress the coordinates of each landmark point.

\parspace
\paragraph{Soft-argmax.}
Since soft-argmax is an important component of our model, we review its formulation in this paragraph. This operator can be defined as a sequence of two steps, where the first one calculates the spatial softmax for pixel $(i, j)$:
\begin{equation}
\Phi(\beta, \mathbf{h}, i, j) = \frac{\exp[\beta\mathbf{h}_{ij}]}{\sum_{k=0}^{W-1}\sum_{l=0}^{H-1} \exp [\beta\mathbf{h}_{kl}]}
\end{equation}

At the next stage, the obtained spatial softmax is multiplied by the expected value of landmark coordinate at every pixel:

\begin{equation}
    \Psi_d (\mathbf{h}) = \sum_{i=0}^{W-1}\sum_{j=0}^{H-1} \mathbf{W}_{ij}^{(d)}\Phi(\beta, \mathbf{h}, i, j),
\end{equation}
where
\begin{equation}
    \mathbf{W}_{ij}^{(x)} = \frac{i}{W},  \mathbf{W}_{ij}^{(y)} = \frac{j}{H}.
\end{equation}

\subsection{Loss function}
We assessed various loss functions for training our model and finalized our choice at wing loss~\cite{feng2018wing} that is closely related to $L_1$ loss. However, in the case of wing loss, the errors in a small vicinity of $0$ -- $(-w, w)$ are better amplified due to the logarithmic nature of the function:

\begin{equation}
    L(y, \hat y) = \left\{
\begin{array}{ll}
      w\log\left(1+\frac{1}{\epsilon}|y-\hat y|\right) & |y-\hat y| < w\\
      |y-\hat y|-C & \textrm{otherwise}\\
\end{array}
\right.,
\end{equation}
where $y$ -- is a ground truth, $\hat y$ -- prediction, ($-w$, $w$) -- range of non-linear part of the loss, $C$ -- constant smoothly linking the linear and non-linear parts.

\subsection{Training techniques}
\paragraph{MixUp}
We use a MixUp technique~\cite{zhang2017mixup} to improve the performance of our method. In particular, MixUp mixes the data inputs $x_1$ and $x_2$, the corresponding keypoint arrays $p_1$ and $p_2$:
\begin{align}
   \lambda &\sim \operatorname{Beta}(\alpha, \alpha) \\
   \lambda^\prime &= \max(\lambda, 1 - \lambda) \label{eqn:lambda_prime} \\
   x^\prime &= \lambda^\prime x_1 + (1 - \lambda^\prime)x_2 \\
   p^\prime &= \lambda^\prime p_1 + (1 - \lambda^\prime)p_2,
\end{align}
 thereby augmenting the dataset with the new interpolated examples. Our implementation of mixup does not differ from the one proposed in the original work\footnote{\url{https://github.com/facebookresearch/mixup-cifar10}} and we do not compute the mixed targets $p^\prime$. In contrast, we rather optimize the following loss function calculated mini-batch-wise:
 \begin{equation}
    L^\prime(x_1, x^\prime, p_1, p_2) = \lambda L(p_1,o_1) + (1-\lambda) L(p_2, o^\prime),
 \end{equation}
where $o_1$ and $o^\prime$ are the outputs of the network for $x_1$ and $x^\prime$, respectively. Here, the points $p_2$ for every point $p_1$ are generated by a simple mini-batch shuffling.
\parspace
\paragraph{Data Augmentation.}
Medical images can vary in appearance due to different data acquisition settings or patient-related anatomical features. To tackle the issue of limited data, we applied the data augmentation. We use geometric and textural augmentations similarly to to the face landmark detection problem~\cite{feng2019mining}. The former included all classes of homographic transformations while the latter included gamma correction, salt and pepper, blur (both median and gaussian) and the addition of a gaussian noise. Interestingly, the homographic transformations were shown effective in improving, for example, self-supervised learning~\cite{laskar2019geometric,melekhov2019dgc}, however only more narrow class of transformation (affine) has been applied to the landmark localization~\cite{feng2019mining} in faces.
\parspace
\paragraph{Transfer learning from low-budget annotations.}
As shown in Fig.~\ref{fig:pipeline}, the problem of localizing the landmarks comprises two stages: identification of the ROI and the actual landmark localization. We previously mentioned the two classes of labels that are needed to train such a pipeline: low-cost ($1-2$ points / image) and high-cost labels ($2+$ points). The low-cost labels can be noisy / inaccurate and are quick to produce, while the high-cost labels require the expert knowledge. In this work, we first train the ROI localization model ($1$ landmark per leg) on the low-cost labels -- knee joint centers (see Fig.~\ref{fig:pipeline}) and then re-use the pre-trained weights from this stage to train the landmark localization model ($16$ landmarks per knee joint).
\section{Experiments}
\subsection{Datasets}
\paragraph{Annotation Process}
For all the following datasets, we applied the same annotations process. Firstly, for all the images in all the datasets we run BoneFinder tool (see Sec.~\ref{sec:baselines}). At the second stage, for every image, a person experienced in knee anatomy and OA manually refine all the landmark points. In Fig.~\ref{fig:pipeline}, we highlight the numbering of the landmarks that we use in this paper. Specifically, we marked the corner landmarks in tibia from $0$ to $8$ and in femur from $9$ to $15$ (lateral to medial). To perform the annotations, we used VGG image annotation tool~\cite{dutta2019vgg}.
\parspace
\paragraph{OAI.}
We trained our model and performed model selection using the images from Osteoarthritis Initiative (OAI) dataset\footnote{\url{https://oai.epi-ucsf.org/datarelease/}}. Roughly $150$ knee joint images per KL grade were sampled to be included into the dataset. The final dataset size comprised $748$ knee joints in total. In the case of the ROI localization, we used a half of the image that corresponded to each knee.
\parspace
\paragraph{Dataset A.}
These data were collected at our  hospital \censored{(Oulu University Hospital, Finland)~\cite{podlipska2016comparison}}, and thus, it comes from a completely different population than OAI (from USA). It includes the images from $81$ subjects, and KL grade-wise the data have the following distribution: 4 knees with KL $0$, $54$ knees with KL $1$, $49$ knees with KL $2$, $29$ knees with KL 3 and $25$ knees with KL $4$. From this dataset, we excluded $1$ knee due to an implant, thereby using $161$ knees for testing of our model.
\parspace
\paragraph{Dataset B.}
This dataset was also acquired from our hospital \censored{(Oulu University Hospital, Finland; ClinicalTrials.gov ID: NCT02937064)} and included originally $107$ subjects. Out of these, 5 knee joints were excluded, thereby making a dataset of $209$ knees ($4$ implants and $1$ due to error during the annotation process). With respect to OA severity, these data had $35$ cases with KL $0$, $84$ with KL $1$, $51$ with KL $2$, $37$ with KL $3$ and $2$ with KL $4$. This dataset was also used solely for testing of our model.
\subsection{Baseline methods}\label{sec:baselines}
We used several baseline methods at the model selection phase and one strong pre-trained baseline method at the test phase. In particular, we used Active Appearance Models~\cite{cootes2001active} and Constrained Local Models~\cite{cristinacce2006feature} with both Image Gradient Orientations (IGO)~\cite{tzimiropoulos2012subspace} and Local Binary Patterns Features (LBP)~\cite{ojala2002multiresolution}. Our implementation is based on the available methods with default hyperparameters from the Menpo library~\cite{alabort2014menpo}.

At the test phase, we used pre-trained RFRV-CLM method~\cite{lindner2013accurate} implemented in BoneFinder tool. Here, the RFRV-CLM model was trained on $500$ images from OAI dataset. However we did not have access to the train data to assess which samples were used for training this method, therefore, we used this tool only for testing on datasets A and B.

\subsection{Implementation Details}
\paragraph{Ablation experiments}
All our ablation experiments were conducted on the same $5$-fold patient-wise cross-validation split stratified by a KL grade to ensure equal distribution of different stages of OA severity. Both ROI and landmark localization models were trained using the same split.

During the training, we used exactly the same hyperparameters for all the experiments. In particular, we used $N=24$ and $d=6$ for our network. The learning rate and the batch size were fixed to $1e-3$ and $16$, respectively. In some of our experiments where the weight decay was used, we set it to $1e-4$. All the models were trained with Adam optimizer~\cite{kingma2014adam}. The pixel spacing for ROI localization was set to $1$ mm and for the landmark localization to $0.3$ mm. We used bi-linear interpolation for image resizing.

All the ablation experiments were conducted solely on landmark localization task and eventually, after selecting the best configuration, we used it for training the ROI localization model due to the similarity of the tasks. We used the ground truth annotations to crop the $140\times140$ mm ROIs around the tibial center (landmark $4$ in Fig.~\ref{fig:pipeline}) to create the data for model selection and training the landmark localization model. In our experiments, we flipped all the left ROI images to look like the right ones, however this strategy was not applied for the ROI localization task.

When performing the fine-tuning of landmark localization model using the pre-trained weights of the ROI localization model, we simply initialized all the layers of the former with the  weights of the latter one. We note here that the last layer was initialized randomly and we did not freeze the pre-trained part for simplicity.

In our experiments, we used PyTorch v$1.1.0$~\cite{paszke2017automatic} on a single Nvidia GTX1080Ti. For data augmentation, we used SOLT library~\cite{tiulpin2019solt}. For training AAM and CLM, we used Menpo~\cite{alabort2014menpo}, as mentioned earlier.
\parspace
\paragraph{Evaluation and Metrics}
To assess the results of our method, we used multiple metrics and evaluation strategies. Firstly, we performed the ablation experiments and used the landmarks $0,8,9,15$ for evaluation of the results (see Fig.~\ref{fig:pipeline}). At the test time, when comparing the performance of the full system, we used an extended set of landmarks for evaluation -- $0,4,8,9,12,15$. The intuition here is to compare the landmark methods on those landmark points that are the most crucial in applications (tibial corners for landmark localization as well as tibial and femoral centers for the ROI localization). Besides, we excluded all the knees with implants from the evaluation.

As as the main metric for comparison, we used Percentage of Correct Keypoints (PCK) $@$~$r$ to compare the landmark localization methods. This metric shows the percentage of points that fall within the neighborhood of a ground truth landmark having the radius $r$ (recall at different precision thresholds). In our experiments, we used $r$ of $1$ mm, $1.5$ mm, $2$ mm and $2.5$ mm for quantitative comparison.

Finally, we also assessed the amount of outliers in the landmark localization task. An outlier was defined as a landmark that do not fall within the $10$ mm neighbourhood of the ground truth landmark. This value was computed for all the landmark points in contrast to PCK.

\subsection{Ablation Study}
\paragraph{Conventional approaches.}
We first investigated the conventional approaches for landmark localization. The benchmarks of AAM and CLM  with IGO and LBP features with default hyperparameters from Menpo~\cite{alabort2014menpo} showed satisfactory results. The best model here was CLM with IGO features (Tab.~\ref{tab:ablation}).

\parspace
\paragraph{Loss Function.}
In the initial experiments with our model we assessed different loss functions ( see Tab.~\ref{tab:ablation}). In particular, we used $L_2$,$L_1$, wing~\cite{feng2018wing} and elastic loss (sum of $L_2$ and $L_1$ losses). Besides, we also utilized a recently introduced general adaptive robust loss with the default hyperparameters~\cite{barron2019general}. Our experiments showed that wing loss with the default hyperparameters as in the original paper ($w=15$ and $C=3$), produces the best results.
\parspace
\paragraph{Effect of Multi-scale Residual Blocks.}
The experiments done for loss functions were conducted using the HMP block. However, it is worth to assess the added value of this block compare to the bottleneck residual block. Tab.~\ref{tab:ablation} demonstrates that the bottleneck residual block ("Wing + regular res. block" of the Table) fell behind of HMP ("Wing loss") in terms of PCK.
\parspace
\paragraph{MixUp vs. Weight Decay}
After observing that the wing loss and HMP block yield the best default configuration, we experimented with various forms of regularization. In this series of experiments, we used our default configuration and applied MixUp with different $\alpha$. Our experiments showed that using MixUp the default configuration and weight decay  degrades the performance (Tab.~\ref{tab:ablation}). However, MixUp itself is also a powerful regularizer, therefore, we conducted the experiments without weight decay (marked as \emph{no wd} in Tab.~\ref{tab:ablation}). Interestingly, setting weight decay to $0$ increases the performance of our model with any $\alpha$. To assess the strength of regularization, we also conducted an experiment with $\alpha=0.75$ (best) and without dropout. We observed that having dropout helps MixUp.
\parspace
\paragraph{CutOut vs. Target Jitter}
Besides MixUp, we tested two other data augmentation techniques -- cutout~\cite{devries2017improved} and noise addition to the ground truth annotations during the training (uniform distribution, $\pm 1$ pixel). We observed that the latter did not improve the results of our configuration with MixUp, however the former helped to lower down the amount of outliers twice while yielding nearly the same localization performance. This configuration had a cutout of $10\%$ of the image. These results are also presented in Tab.~\ref{tab:ablation}.

\begin{figure}
    \centering
    \subfloat[]{\includegraphics[width=0.4\linewidth]{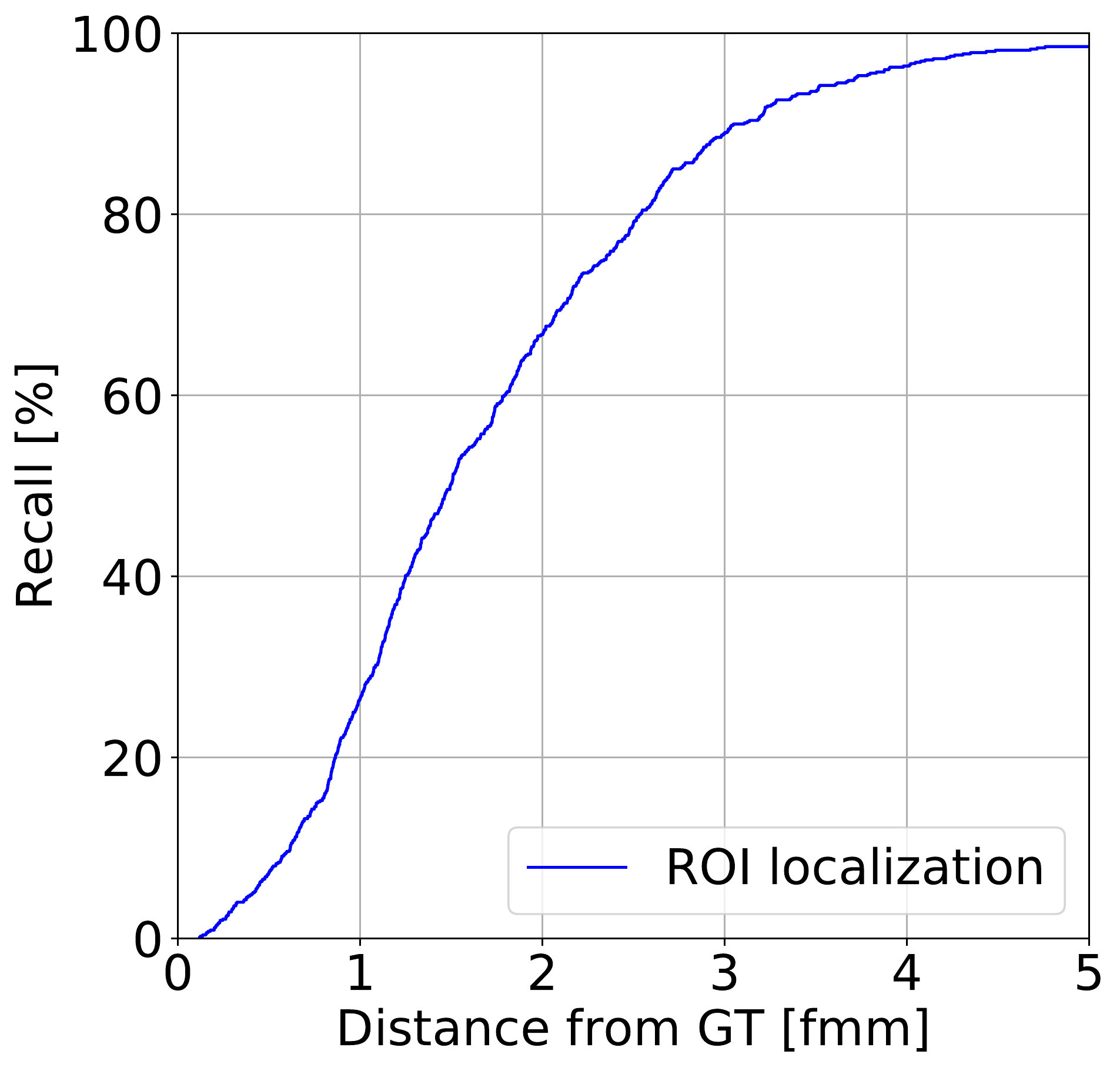}}\hfill
    \subfloat[]{\includegraphics[width=0.4\linewidth]{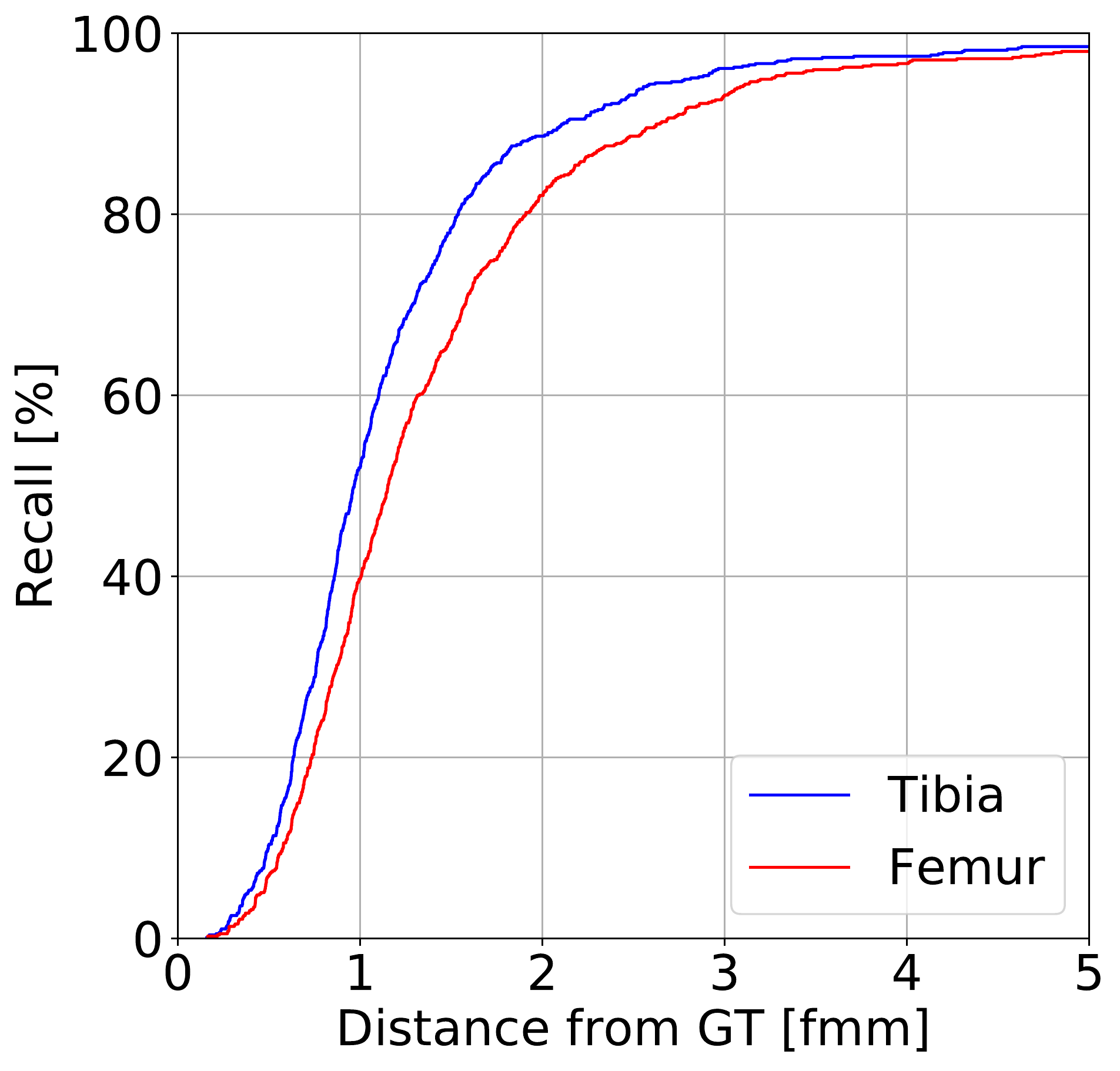}}
    \vspace{0.1cm}
    \caption{Cumulative plots reflecting the performance of ROI (a) and landmark (b) localization methods on cross-validation. ROI localization was assessed at the pixel spacing of $1$ mm and the landmark localization at $0.3$ mm, respectively. GT indicates ground truth.}
    \label{fig:cv-perf}
\end{figure}

\paragraph{Transfer Learning from Low-cost Labels.}
At the final stage of our experiments, we used the best configuration that included the wing loss, MixUp with $\alpha =0.75$, weight decay of $0$ and $10\%$ cutout to train the ROI localization model. Essentially, both of these methods are landmark localization approaches, therefore, in our cross-validation experiments, we also assessed the performance of ROI localization using PCK. In our experiments, we found that pre-training of the landmark localization model on the ROI localization task significantly increases the performance of the former (see the last row of Tab.~\ref{tab:ablation}). The performance of both these models on cross-validation is presented in Fig.~\ref{fig:cv-perf}. Quantitatively, ROI localization model yielded PCK of $26.60\%$, $50.27\%$, $66.71\%$, $79.14\%$ at $1$ mm, $1.5$ mm, $2$ mm and $2.5$ mm thresholds, respectively and had $0.13\%$ outliers.

\begin{table*}[ht!]
\centering
{
\footnotesize
\begin{tabular}{ccccccc}
\toprule
\textbf{Setting} & \textbf{1 mm} & \textbf{1.5 mm} & \textbf{2 mm} & \textbf{2.5 mm} & \textbf{\% out}\\
\midrule
AAM (IGO~\cite{tzimiropoulos2012subspace}) & $7.29 \pm 4.06$ & $17.18 \pm 5.39$ & $28.07 \pm 5.29$ & $39.51 \pm 6.33$ & $7.49$\\
AAM (LBP~\cite{ojala2002multiresolution}) & $2.41 \pm 0.19$ & $8.02 \pm 1.13$ & $15.17 \pm 3.12$ & $24.33 \pm 4.73$ & $9.22$\\
CLM (IGO~\cite{tzimiropoulos2012subspace}) & $24.53 \pm 3.31$ & $39.84 \pm 4.92$ & $50.60 \pm 3.69$ & $61.43 \pm 4.25$ & $3.61$\\
CLM (LBP~\cite{ojala2002multiresolution}) & $2.67 \pm 1.51$ & $10.03 \pm 3.21$ & $18.65 \pm 5.77$ & $28.81 \pm 5.58$ & $9.36$\\

\midrule
L2 loss& $0.00 \pm 0.00$ & $0.00 \pm 0.00$ & $0.07 \pm 0.09$ & $0.07 \pm 0.09$ & $92.78$\\
L1 loss &  $17.45 \pm 5.20$ & $45.45 \pm 5.48$ & $66.11 \pm 5.39$ & $80.08 \pm 3.78$ & $2.67$\\
Robust loss~\cite{barron2019general} & $13.97 \pm 0.47$ & $35.83 \pm 1.70$ & $57.35 \pm 1.89$ & $72.06 \pm 1.89$ & $4.68$\\
Elastic loss & $4.14 \pm 3.40$ & $13.97 \pm 7.66$ & $27.21 \pm 9.74$ & $41.58 \pm 10.59$ & $9.36$\\
Wing loss~\cite{feng2018wing} & $31.68 \pm 5.10$ & $61.83 \pm 7.09$ & $78.68 \pm 5.58$ & $87.50 \pm 3.31$ & $2.14$\\

\midrule
Wing + regular res. block & $25.74 \pm 3.31$ & $55.48 \pm 3.97$ & $73.46 \pm 3.69$ & $83.82 \pm 3.03$ & $2.67$\\

\midrule
Wing + mixup $\alpha=0.1$ & $27.54 \pm 0.19$ & $58.42 \pm 1.70$ & $77.21 \pm 1.42$ & $87.17 \pm 0.57$ & $2.27$\\
Wing + mixup $\alpha=0.2$ & $29.88 \pm 4.25$ & $58.96 \pm 2.84$ & $78.07 \pm 6.05$ & $86.16 \pm 3.50$ & $2.94$\\
Wing + mixup $\alpha=0.5$ & $29.61 \pm 1.42$ & $59.36 \pm 3.03$ & $77.81 \pm 3.78$ & $86.30 \pm 2.55$ & $2.67$\\
Wing + mixip $\alpha=0.75$ & $30.75 \pm 3.40$ & $59.63 \pm 4.92$ & $77.07 \pm 5.20$ & $86.36 \pm 2.84$ & $3.48$\\

Wing + mixup $\alpha=0.1$ (no wd) & $34.89 \pm 5.29$ & $63.64 \pm 7.56$ & $81.15 \pm 5.48$ & $89.24 \pm 3.12$ & $1.47$\\
Wing + mixup $\alpha=0.2$ (no wd) & $35.16 \pm 5.86$ & $64.17 \pm 7.00$ & $82.15 \pm 5.58$ & $89.91 \pm 4.25$ & $1.34$\\
Wing + mixup $\alpha=0.5$ (no wd) & $36.30 \pm 6.33$ & $65.04 \pm 6.33$ & $81.82 \pm 4.16$ & $89.91 \pm 2.55$ & $1.47$\\
Wing + mixup $\alpha=0.75$ (no wd) & $37.97 \pm 5.48$ & $67.45 \pm 4.25$ & $82.02 \pm 1.80$ & $90.51 \pm 0.95$ & $1.60$\\
\midrule
Wing + mixup $\alpha=0.75$ (no wd, no dropout) & $37.10 \pm 5.39$ & $65.64 \pm 3.97$ & $81.75 \pm 4.44$ & $89.30 \pm 3.21$ & $1.47$\\
\midrule
Wing + mixup $\alpha=0.75$ + jitter (no wd) & $36.63 \pm 4.16$ & $65.98 \pm 5.58$ & $83.09 \pm 3.88$ & $90.84 \pm 3.31$ & $1.60$\\

\midrule
Wing + mixup $\alpha=0.75$ + cutout 5\% (no wd)& $34.96 \pm 3.69$ & $63.30 \pm 6.14$ & $80.15 \pm 4.06$ & $89.30 \pm 1.32$ & $1.07$\\
Wing + mixup $\alpha=0.75$ +  cutout 10\% (no wd)& $37.83 \pm 4.35$ & $65.78 \pm 4.35$ & $81.35 \pm 3.50$ & $90.24 \pm 1.51$ & $\mathbf{0.53}$\\
Wing + mixup $\alpha=0.75$ + cutout 25\% (no wd)& $35.56 \pm 3.97$ & $62.50 \pm 5.01$ & $80.01 \pm 4.06$ & $88.50 \pm 2.84$ & $0.94$\\

\midrule
Wing + mixup $\alpha=0.75$ + cutout 10\% (no wd, finetune) & $\mathbf{45.92 \pm 8.79}$ & $\mathbf{72.39 \pm 8.60}$ & $\mathbf{85.36 \pm 4.63}$ & $\mathbf{90.91 \pm 3.21}$ & $1.34$\\
\bottomrule
\end{tabular}
\vspace{0.1cm}
\caption{Results of the model selection for high-cost annotations on the OAI dataset. The values of PCK/recall (\%) at different precision are shown as average and standard deviation for the landmarks $0$, $8$, $9$, $15$, while the amount of outliers is calculated for all the landmarks. The comparison is done at $0.3$ mm image resolution (pixel spacing). Best results are highlighted in bold.}\label{tab:ablation}

}
\end{table*}

\subsection{Test datasets}
\paragraph{Testing on the full datasets}
Testing of our model was conducted on datasets A and B, respectively.  We provide the quantitative results in Tab.~\ref{tab:test}.
In this table, we present two versions of our pipeline, one is a single stage, where the landmark localization follows directly after the ROI localization step, and also a two-stage pipeline that includes ROI localization as a first step, initial inference of the landmark points as a second step, and re-centering of the ROI to the predicted tibial center and a second pass of landmark localization model as a third step. 


\begin{table*}[ht!]
\centering
{
\footnotesize
\begin{tabular}{lllllll}
\toprule
\multirow{2}{*}{\textbf{Dataset}}              & \multicolumn{1}{c}{\multirow{2}{*}{\textbf{Method}}} & \multicolumn{4}{c}{\textbf{Precision}}                                                                        & \multicolumn{1}{c}{\multirow{2}{*}{\textbf{\% out}}} \\ \cmidrule(lr){3-6}
                                               & \multicolumn{1}{c}{} & \multicolumn{1}{c}{1 mm} & \multicolumn{1}{c}{1.5 mm} & \multicolumn{1}{c}{2 mm} & \multicolumn{1}{c}{2.5 mm} & \multicolumn{1}{c}{}                                 \\
\midrule
\multirow{3}{*}{A} & BoneFinder~\cite{lindner2013accurate} & $\mathbf{48.45 \pm 2.64}$ & $\mathbf{59.63 \pm 3.51}$ & $78.26 \pm 7.03$ & $89.13 \pm 3.95$ & $\mathbf{0.00}$ \\
                                               & Ours 1-stage & $12.73 \pm 2.20$ & $46.89 \pm 5.71$ & $78.57 \pm 1.32$ & $90.99 \pm 1.32$ & $1.24$ \\
                                               & Ours 2-stage & $14.60 \pm 4.83$ & $47.52 \pm 2.20$ & $\mathbf{78.88 \pm 0.88}$ & $\mathbf{93.48 \pm 0.44}$ & $0.62$ \\
\midrule
\multicolumn{1}{l}{\multirow{3}{*}{B}} & BoneFinder~\cite{lindner2013accurate} & $2.87 \pm 3.38$ & $13.64 \pm 10.49$ & $43.78 \pm 21.31$ & $68.90 \pm 20.98$ & $\mathbf{0.00}$\\
\multicolumn{1}{l}{}  & Ours 1-stage & $9.33 \pm 1.01$ & $42.58 \pm 1.35$ & $74.40 \pm 1.69$ & $91.63 \pm 1.69$ & $0.48$ \\
\multicolumn{1}{l}{}  & Ours 2-stage  & $\mathbf{11.24 \pm 0.34}$ & $\mathbf{44.98 \pm 0.68}$ & $\mathbf{75.12 \pm 2.71}$ & $\mathbf{92.11 \pm 0.34}$ & $0.48$ \\
\bottomrule
\end{tabular}
\vspace{0.1cm}
\caption{Test set results and comparison to the state-of-the-art method (RFRV-CLM-based BoneFinder tool) by Lindner~\emph{et al}.~\cite{lindner2013accurate}. Reported percentage of outliers is calculated for \emph{all} landmarks, while the PCK/recall values (\%) are calculated as the average for the landmarks $0$, $4$, $8$, $9$, $12$, and $15$.  Best results per dataset are highlighted in bold. It should be noted that BoneFinder operated with the full image resolution while our method performed ROI localization at $1$ mm and landmark localization at $0.3$ mm resolutions, respectively.}\label{tab:test}
}
\end{table*}

\begin{figure*}
    \centering
    \subfloat[Dataset A (no OA)]{\includegraphics[width=0.2\textwidth]{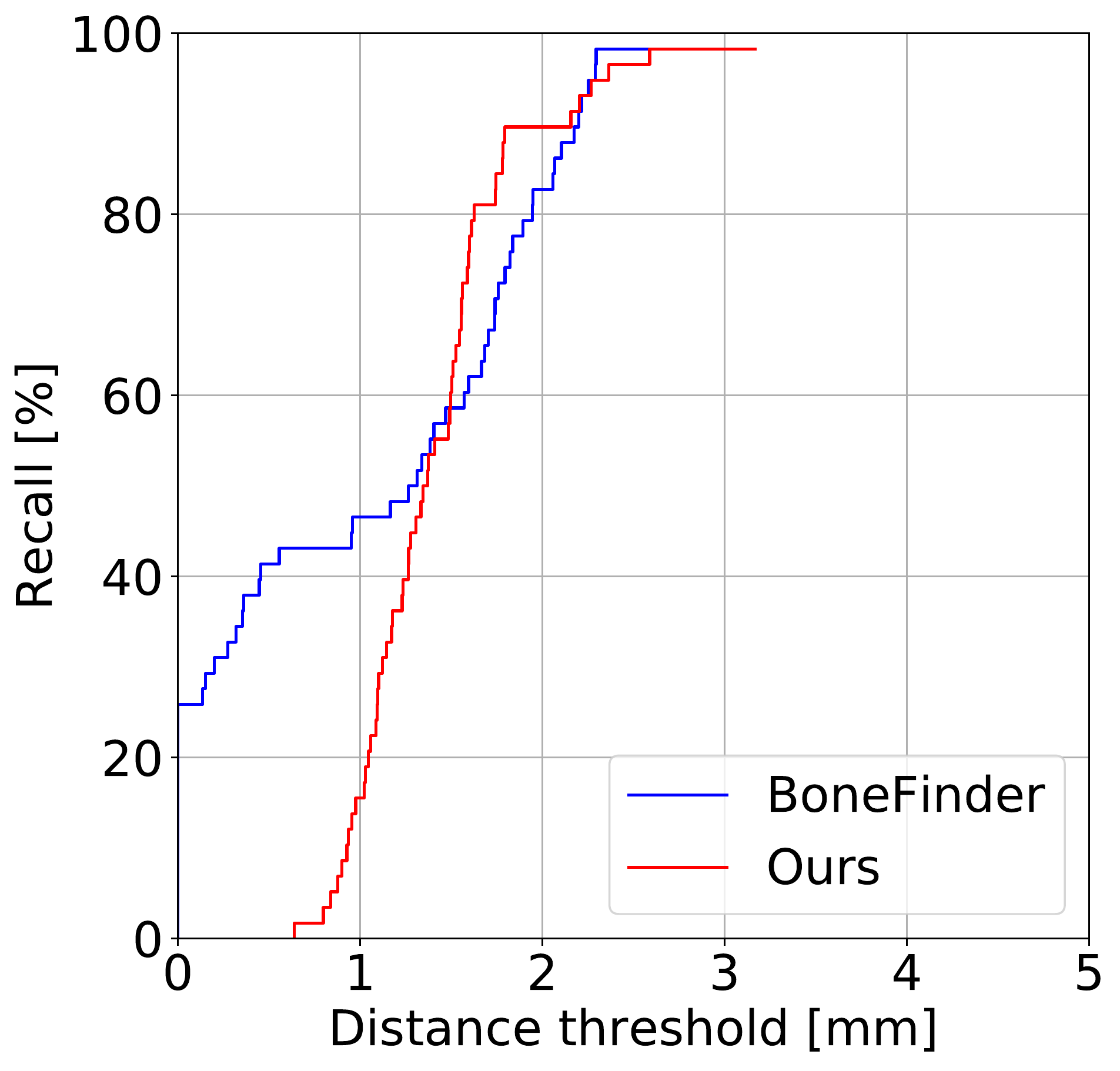}}
    \hfill
    \subfloat[Dataset A (OA)]{\includegraphics[width=0.2\textwidth]{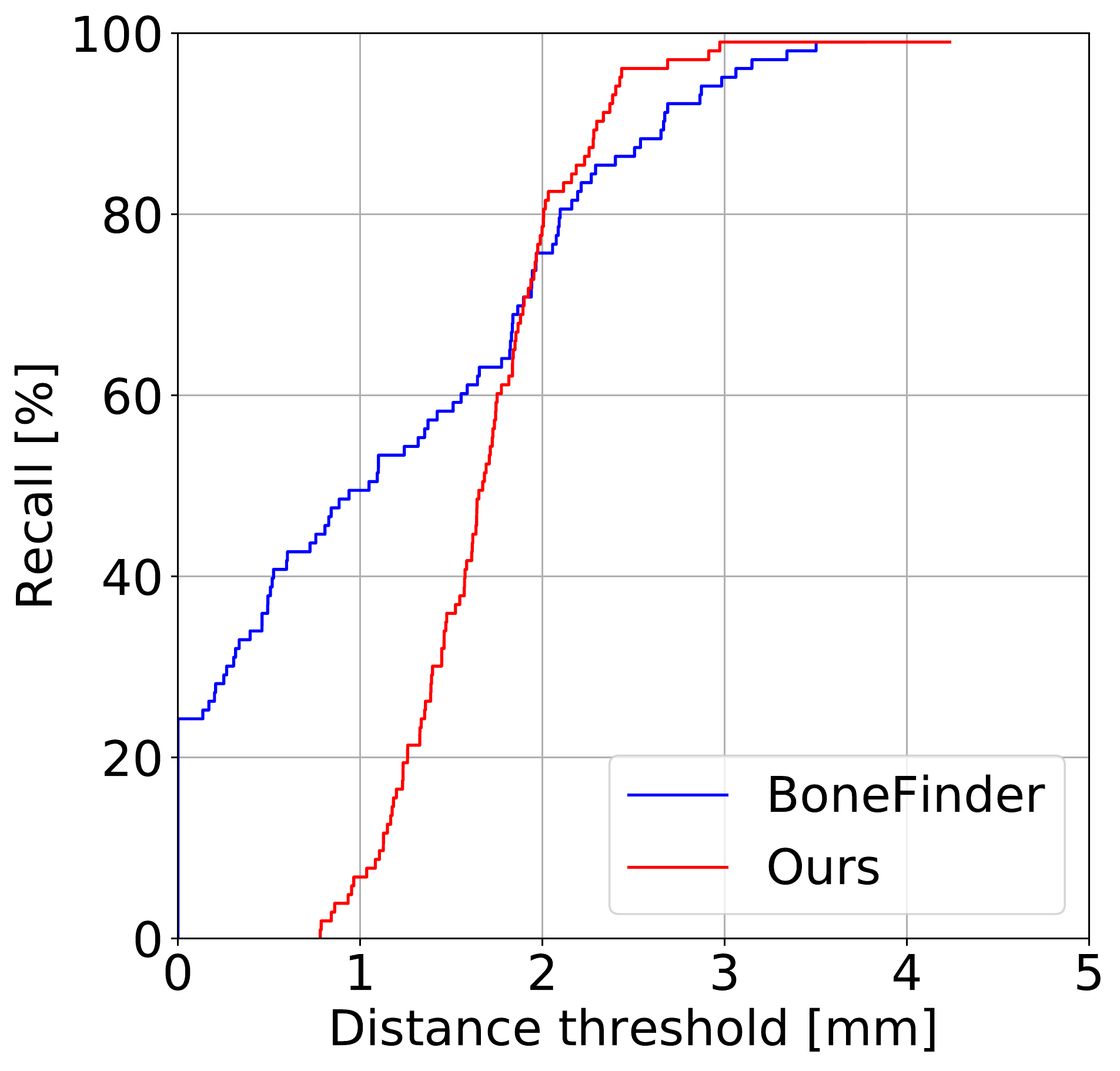}}
    \hfill
    \subfloat[Dataset B (no OA)]{\includegraphics[width=0.2\textwidth]{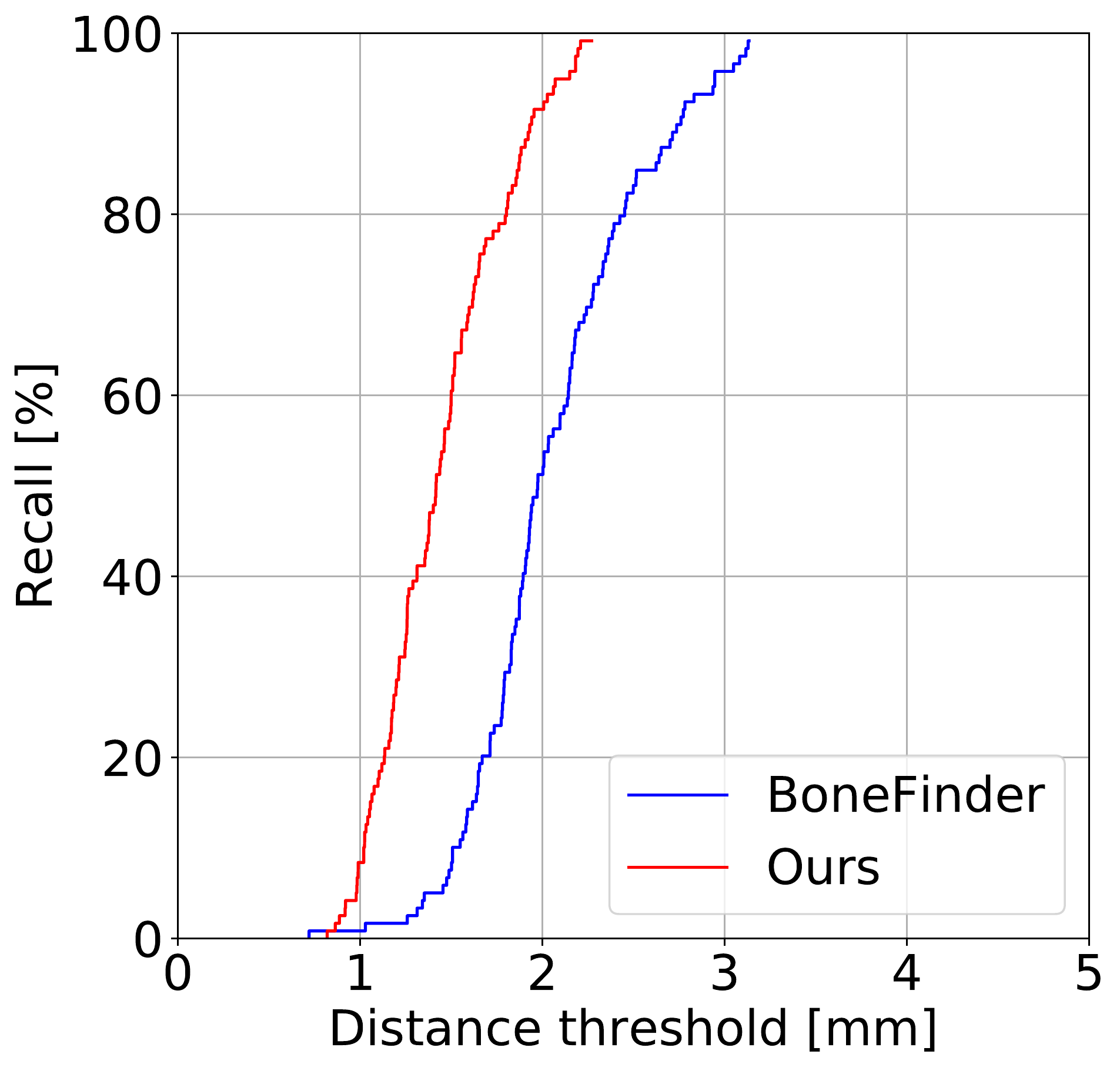}}
    \hfill
    \subfloat[Dataset B (OA)]{\includegraphics[width=0.2\textwidth]{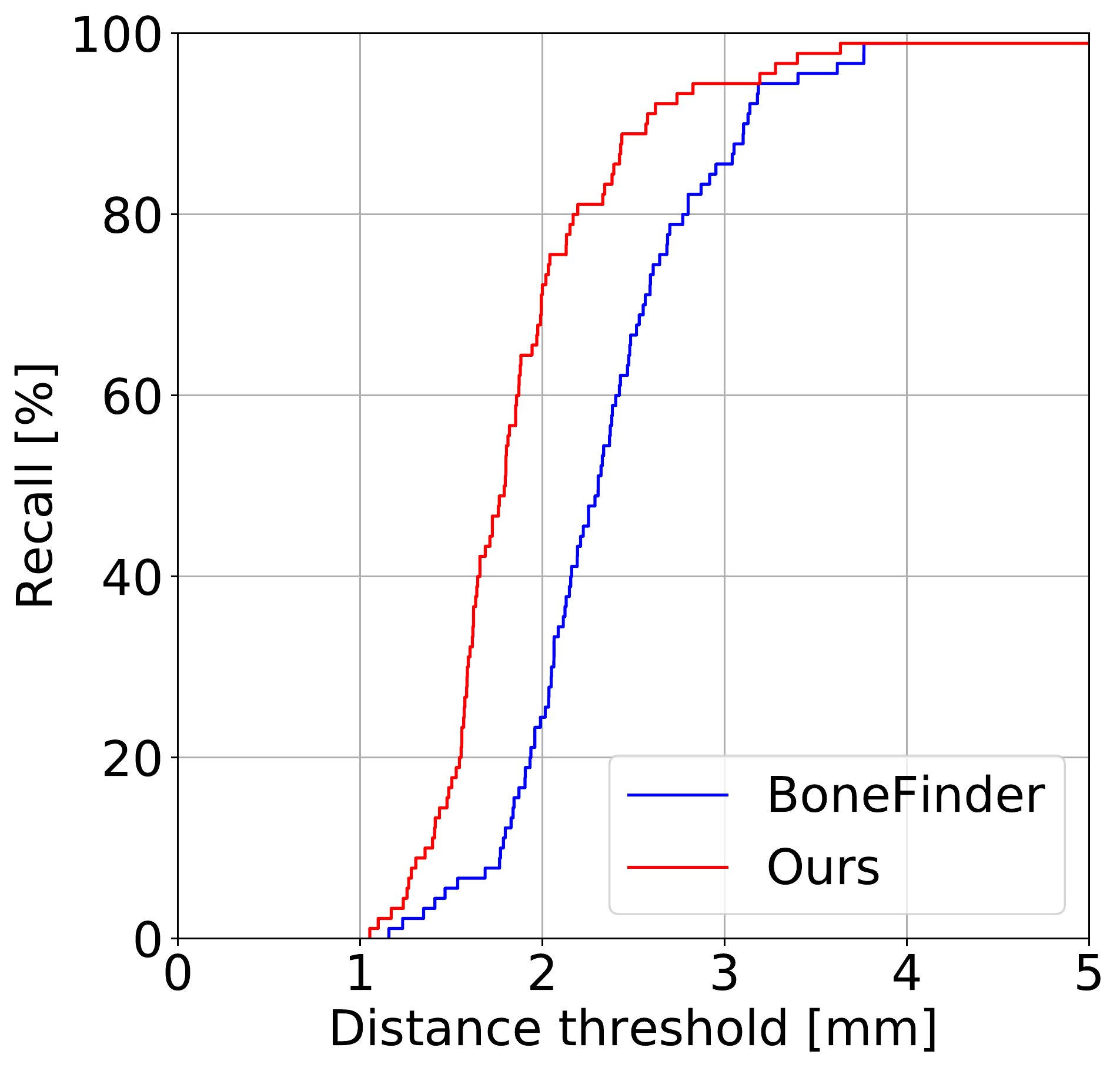}}
    \vspace{0.1cm}
    \caption{Cumulative distribution plots of localization errors for our two-stage method and BoneFinder~\cite{lindner2013accurate,lindner2014robust} for cases with and without radiographic OA on datasets A and B, respectively.}
    \label{fig:curves-test-oa}
\end{figure*}

\begin{figure*}
    \centering
    \subfloat{\includegraphics[width=0.25\textwidth]{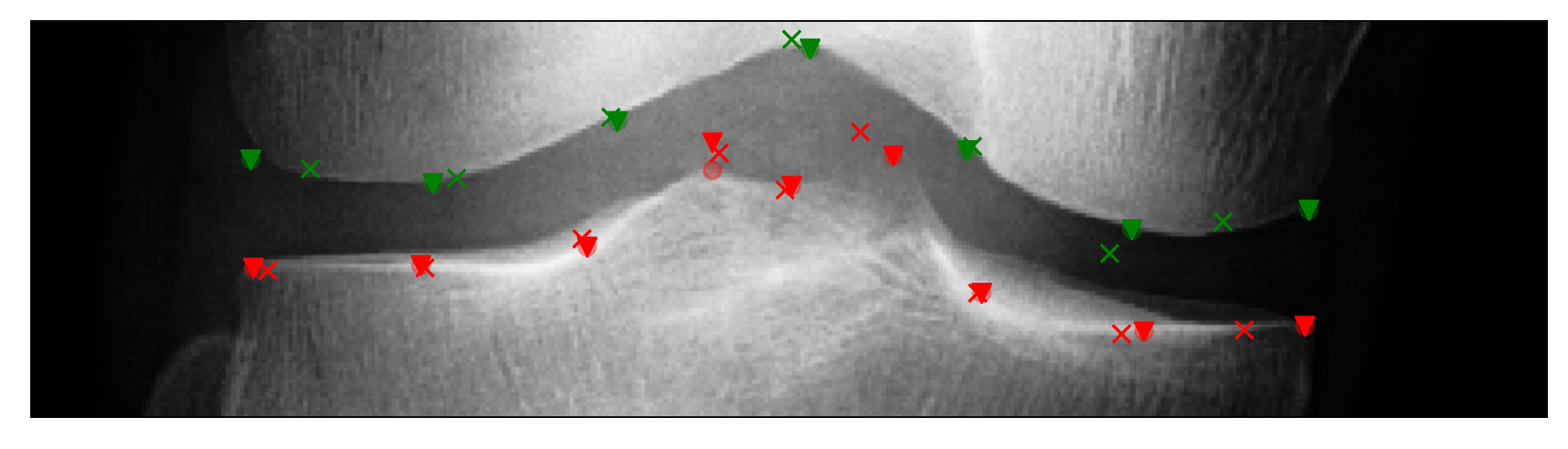}}
    \subfloat{\includegraphics[width=0.25\textwidth]{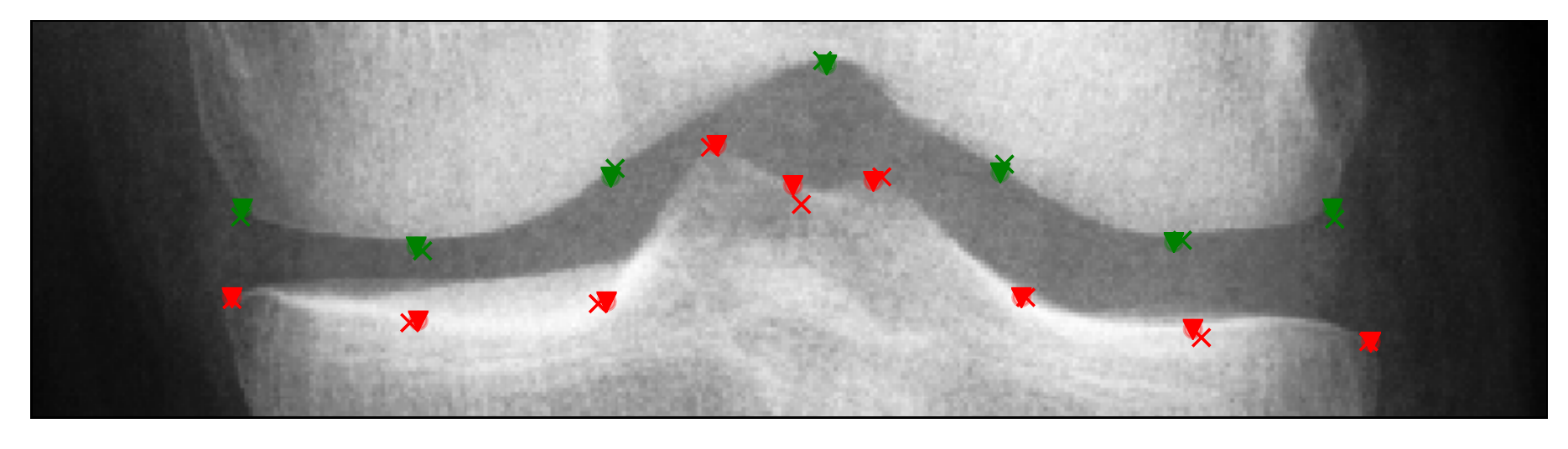}}
    \subfloat{\includegraphics[width=0.25\textwidth]{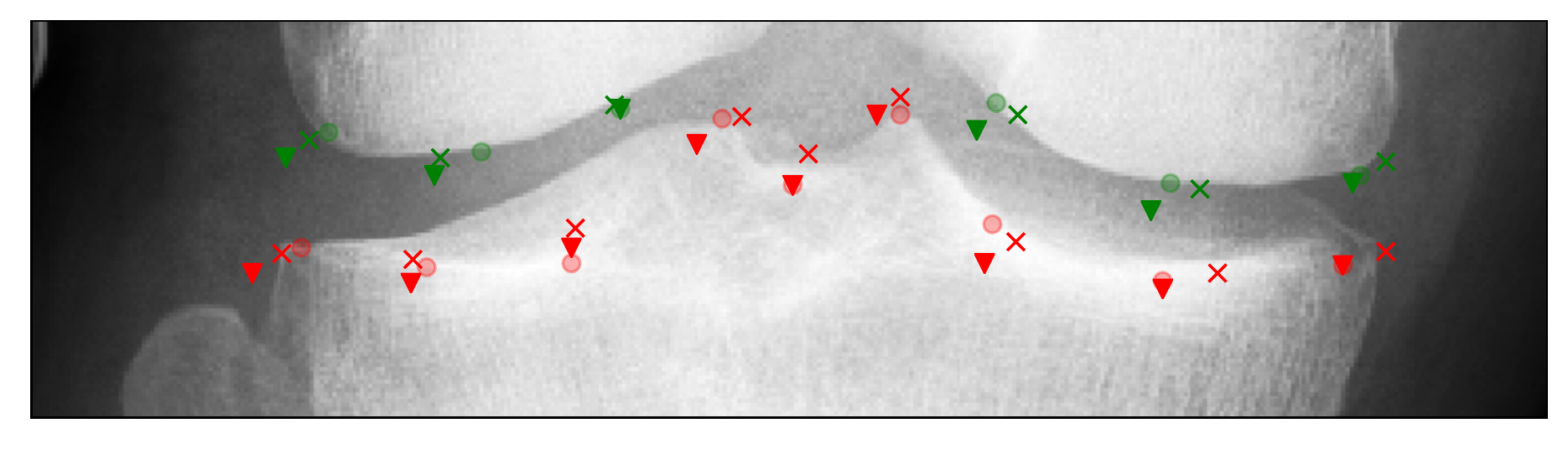}}
    \subfloat{\includegraphics[width=0.25\textwidth]{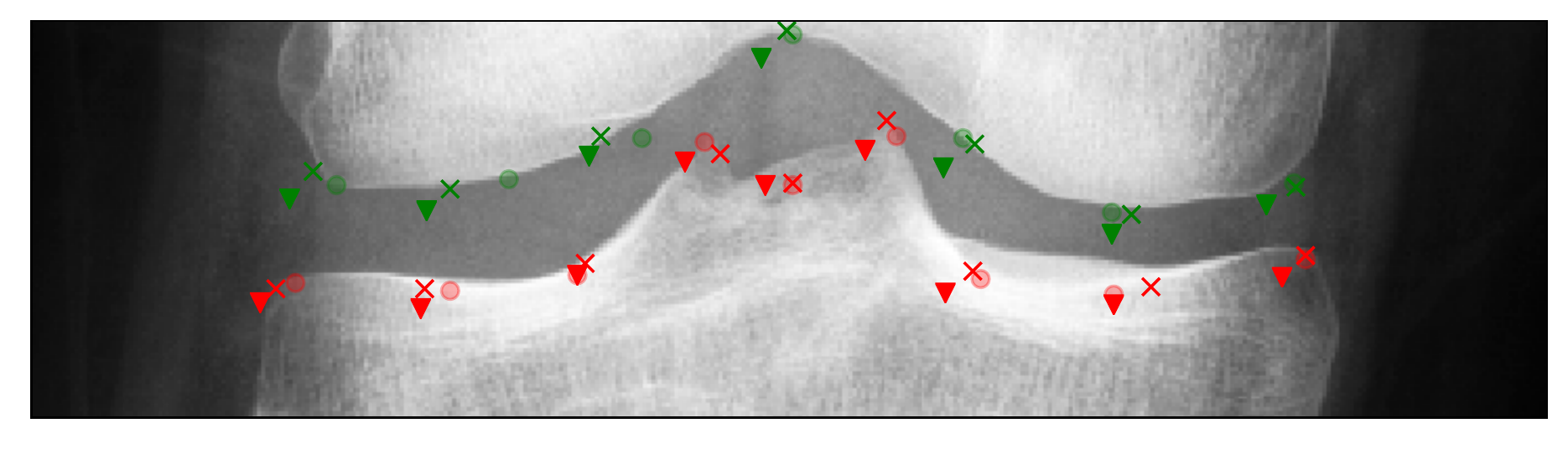}}
    \vspace{-4mm}
    \subfloat{\includegraphics[width=0.25\textwidth]{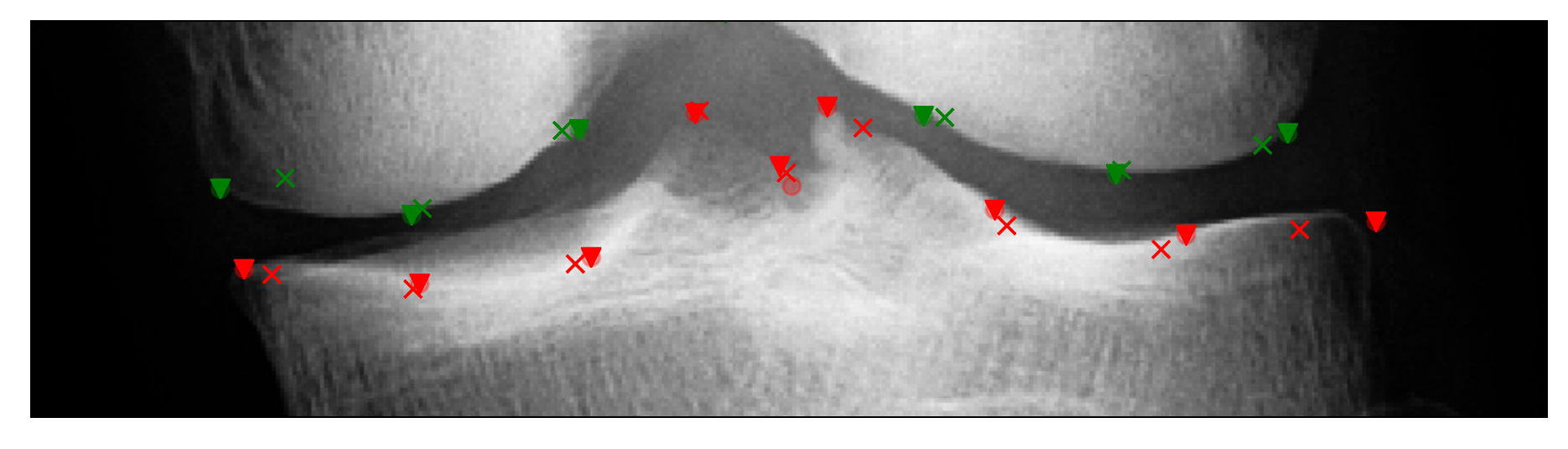}}
    \subfloat{\includegraphics[width=0.25\textwidth]{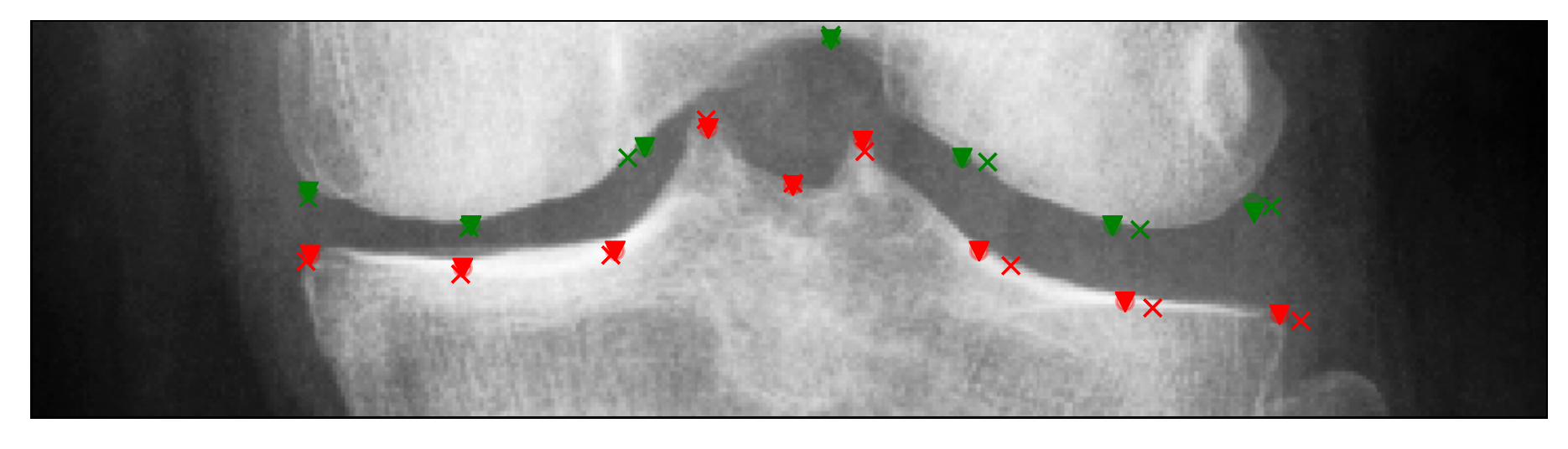}}
    \subfloat{\includegraphics[width=0.25\textwidth]{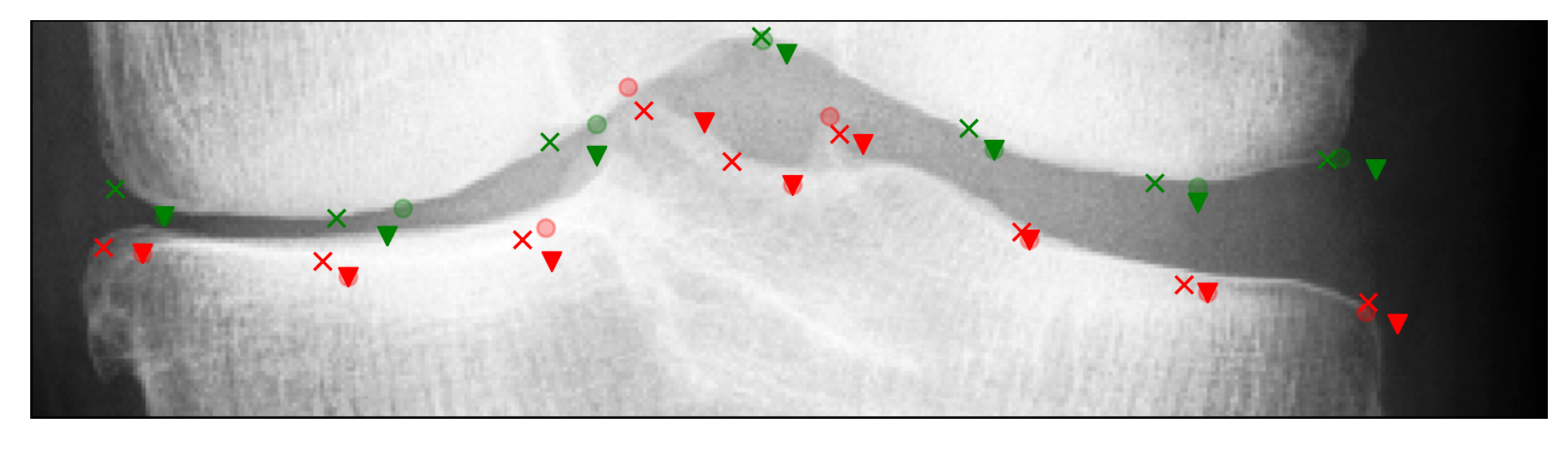}}
    \subfloat{\includegraphics[width=0.25\textwidth]{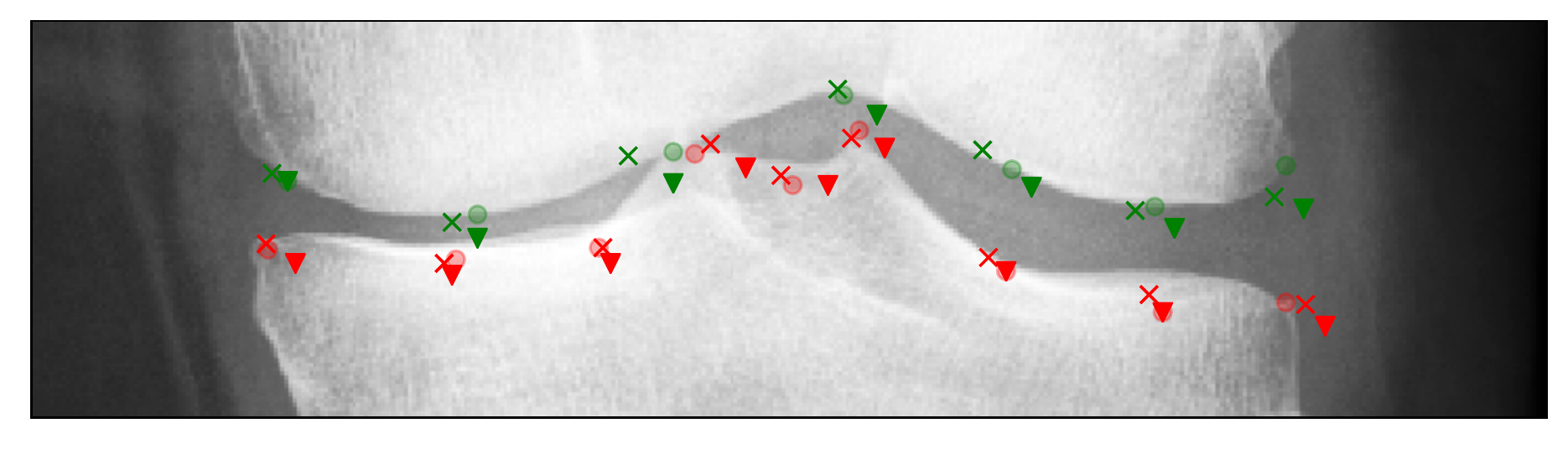}}
    \vspace{-4mm}\setcounter{subfigure}{0}
    \subfloat[Dataset A (worst)]{\includegraphics[width=0.25\textwidth]{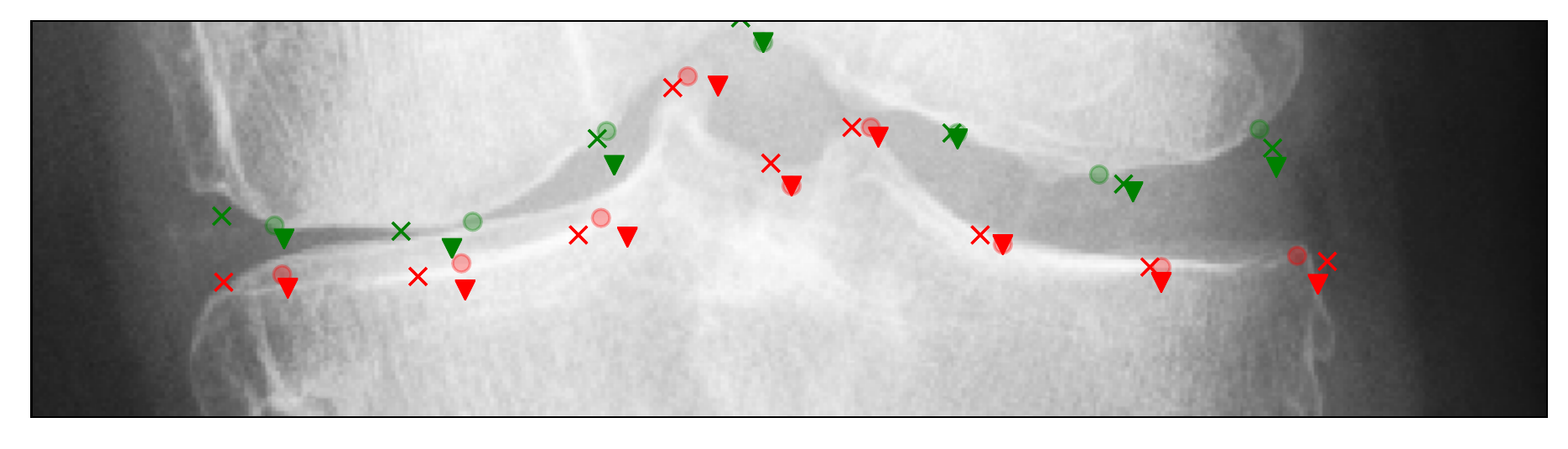}}
    \subfloat[Dataset A (best)]{\includegraphics[width=0.25\textwidth]{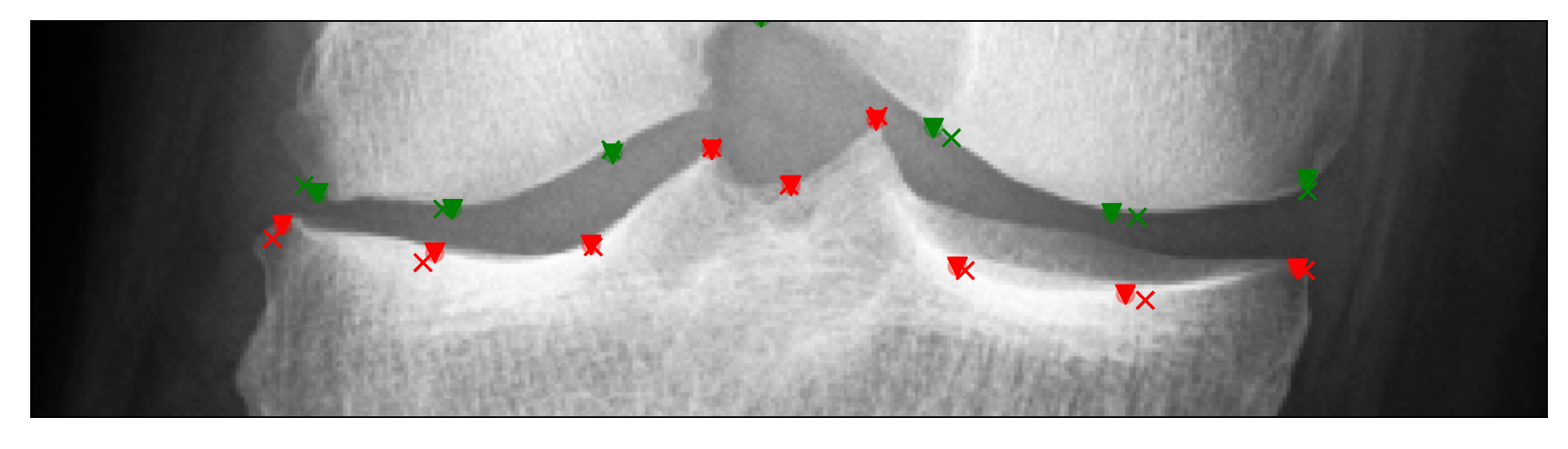}}
    \subfloat[Dataset B (worst)]{\includegraphics[width=0.25\textwidth]{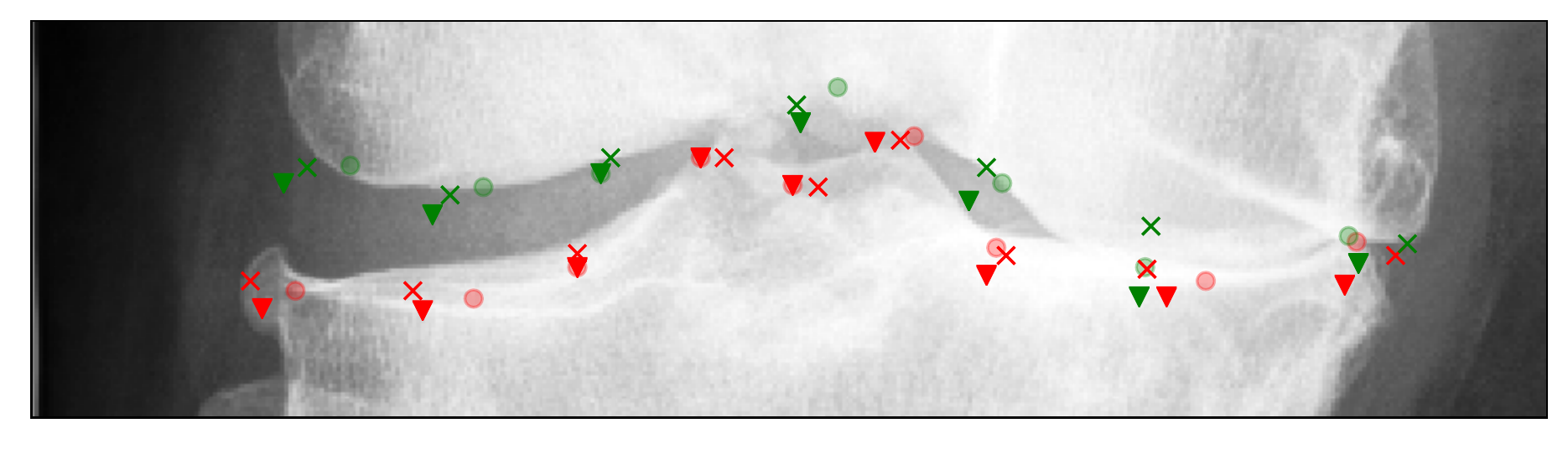}}
    \subfloat[Dataset B (best)]{\includegraphics[width=0.25\textwidth]{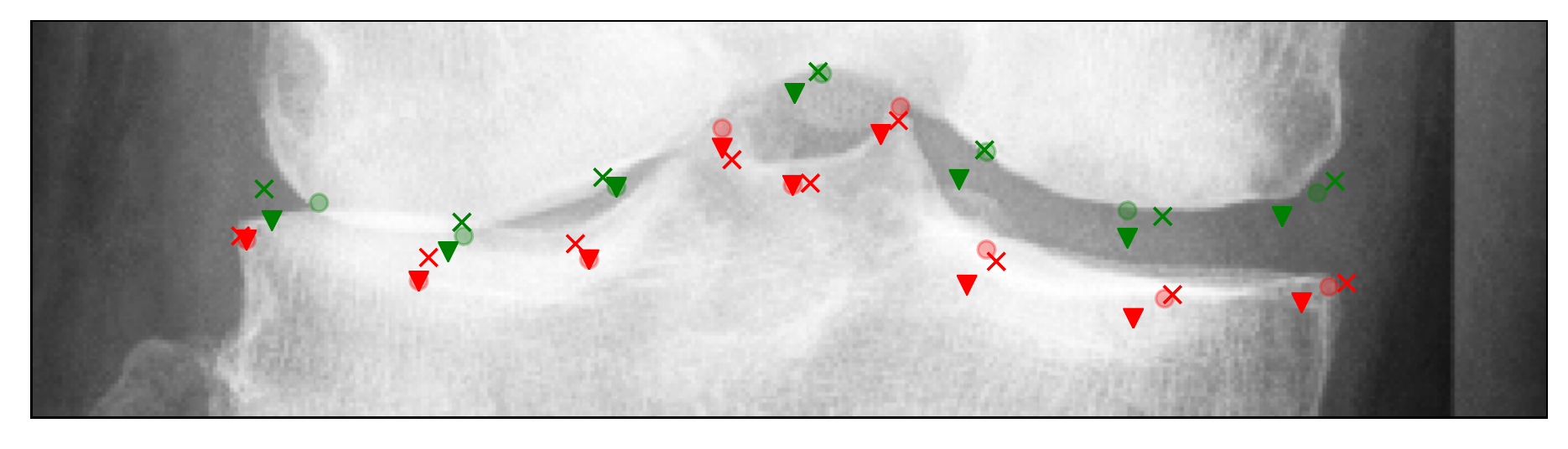}}
    \vspace{0.1cm}
    \caption{Examples of predictions on datasets A and B (worst and best cases). We visualized ground truth landmarks as circles. Predictions made by our method are shown using crosses and predictions made by BoneFinder are shown using triangles. Red and green show the landmarks for tibia and femur, respectively. Best and worst cases were selected based on the average total error of \emph{our method} per group. The width of every example is $115$ mm. The first row contains examples having KL $0$ or $1$, the second row contains examples with KL $2$ and the third row with KL $3$.}
    \label{fig:examples}
\end{figure*}

\parspace
\paragraph{Testing with Respect to the presence of Radiographic Osteoarthritis}
To better understand the behaviour of our model on the test datasets, we investigated the performance of our 2-stage pipeline and BoneFinder for cases having KL $<2$ and KL $\geq2$, respectively. These results are presented in Fig.~\ref{fig:curves-test-oa}. Our method performs on par with BoneFinder for Dataset A and even exceeds its localization performance for precision thresholds above $2$ mm for radiograhic OA. In Dataset B, on average, our method performs better than BoneFinder when both methods are benchmarked for both non-OA and OA cases. To provide better insights into the performance of our method for different stages of OA severity, we show examples of landmark localization done by our method, BoneFinder and manually (Fig.~\ref{fig:examples}).

\section{Conclusions}
In this paper, the problem of anatomical landmark localization in knee radiographs was addressed. We proposed a new method that combines the power of latest advances in facial landmark localization and pose estimation that allowed us to accurately localize the landmarks on the unseen data.

Compare to the current state-of-the-art~\cite{lindner2014robust,lindner2013accurate}, our method generalized better to the unseen test datasets that had completely different acquisition settings and patient populations. Consequently, these results suggest that our new method may be easily applicable to various tasks in clinical and research settings.

Our study has still some limitations. Firstly, the comparison with BoneFinder should ideally be conducted when it is trained on the same $0.3$ mm resolution data with the same KL grade-wise stratification, or at full image resolution. However, we did not have access to the training code of BoneFinder, thereby, leaving more systematic comparison to future studies. Another limitation of this study is the ground truth annotation process. Specifically, we used BoneFinder to pre-annotate the landmark for all the images in both train and test sets. In theory, this might give an advantage to BoneFinder compared to our method. On the other hand, all the landmarks were still manually refined, which should decrease this advantage.

The core methodological novelties of the study were in adapting the MixUp, soft-argmax layer and transfer learning from low-cost annotations for training our model. We think that the latter has  applications in other, even non-medical domains, such as human pose estimation and facial landmark localization. It was shown that compared to RFRV-CLM, Deep Learning methods scale with the amount of training data, and therefore, we also expect our method to yield even better results when it is trained on a larger datasets~\cite{davison2018landmark}. Besides, we also expect semi-supervised learning~\cite{honari2018improving} to help in this task.

To summarize, we developed a robust method for anatomical landmark localization that has potential to scale with the amount of training data and be applied in the other domains. Our source codes and the annotations made for OAI dataset will be made publicly available.

\section{Acknowledgements}
\censored{This study was supported by KAUTE foundation, Infotech Oulu, University of Oulu strategic funding and Sigrid Juselius Foundation.}

The OAI is a public-private partnership comprised of five contracts (N01- AR-2-2258; N01-AR-2-2259; N01-AR-2- 2260; N01-AR-2-2261; N01-AR-2-2262) funded by the National Institutes of Health, a branch of the Department of Health and Human Services, and conducted by the OAI Study Investigators. Private funding partners include Merck Research Laboratories; Novartis Pharmaceuticals Corporation, GlaxoSmithKline; and Pfizer, Inc. Private sector funding for the OAI is managed by the Foundation for the National Institutes of Health.

Development and maintenance of VGG Image Annotator (VIA) is supported by EPSRC programme grant Seebibyte: Visual Search for the Era of Big Data (EP/M013774/1).

\censored{We thank Dr. Claudia Lindner for providing BoneFinder.}

{\small
\bibliographystyle{ieee}
\bibliography{egbib}
}
\end{document}